\documentclass[a4paper]{article}

\usepackage[english]{babel}
\usepackage[utf8x]{inputenc}
\usepackage[T1]{fontenc}

\usepackage[a4paper,top=3cm,bottom=2cm,left=3cm,right=3cm,marginparwidth=1.75cm]{geometry}

\usepackage{lmodern}
\usepackage{amsmath}
\usepackage{amssymb}
\usepackage{graphicx}
\usepackage{algorithm}
\usepackage{algpseudocode}
\usepackage{algorithmicx}
\usepackage{courier}
\usepackage{varwidth}
\usepackage{subfigure}
\usepackage{epigraph}
\usepackage[colorlinks=true, allcolors=blue]{hyperref}
\usepackage{url}

\usepackage{longtable}  
\usepackage{pdflscape}  
\usepackage{booktabs}  

\usepackage{natbib}

\usepackage[frozencache]{minted}
\usepackage{authblk}  

\newcommand{\ie}{\emph{i.e.},~}
\newcommand{\eg}{\emph{e.g.},~}
\newcommand{\etc}{\emph{etc.}}

\newcommand{\code}[1]{\texttt{#1}}
\newcommand{\literal}[1]{``\texttt{#1}''}

\newcommand{\vocab}{V}

\newcommand{\bc}{\boldsymbol{c}}

\newcommand{\dir}[1]{\emph{\code{#1}}}
\newcommand{\entity}[1]{\emph{\code{#1}}}
\newcommand{\predicate}[1]{\code{#1}}
\newcommand{\cmd}[1]{\textbf{\small{\code{#1}}}}

\newcommand{\domain}[1]{\mathbb{#1}}
\newcommand{\realdomain}{\domain{R}}

\usepackage[export]{adjustbox}  

\begin{document}

\title{TextWorld: A Learning Environment for Text-based Games}
\author[1]{Marc-Alexandre C\^ot\'e\thanks{macote@microsoft.com}}
\author[2]{\'Akos K\'ad\'ar}
\author[1]{Xingdi Yuan}
\author[3]{Ben Kybartas}
\author[1]{Tavian Barnes}
\author[1]{Emery Fine}
\author[1]{James Moore}
\author[1]{Matthew Hausknecht}
\author[3]{Ruo Yu Tao}
\author[1]{Layla El Asri}
\author[1]{Mahmoud Adada}
\author[1]{Wendy Tay}
\author[1]{Adam Trischler}
\affil[1]{Microsoft Research}
\affil[2]{Tilburg University}
\affil[3]{McGill University}



\date{\vspace{-5ex}}  

\maketitle

\epigraph{The limits of my language mean the limits of my world.}{\textit{Ludwig Wittgenstein}}

\begin{abstract}
We introduce TextWorld, a sandbox learning environment for the training and evaluation of RL agents on text-based games.
TextWorld is a Python library that handles interactive play-through of text games, as well as backend functions like state tracking and reward assignment.
It comes with a curated list of games whose features and challenges we have analyzed.
More significantly, it enables users to handcraft or automatically generate new games.
Its generative mechanisms give precise control over the difficulty, scope, and language of constructed games, and can be used to relax challenges inherent to commercial text games like partial observability and sparse rewards.
By generating sets of varied but similar games, TextWorld can also be used to study generalization and transfer learning.
We cast text-based games in the Reinforcement Learning formalism, use our framework to develop a set of benchmark games, and evaluate several baseline agents on this set and the curated list.

\end{abstract}



\section{Introduction}
    \label{sect:introduction}

    Text-based games are complex, interactive simulations in which text describes the game state and players make progress by entering text commands.
    They are fertile ground for language-focused machine learning research.
    In addition to language understanding, successful play requires skills like long-term memory and planning, exploration (trial and error), and common sense.

\begin{figure}
        \centering
        \includegraphics[width=0.8\textwidth]{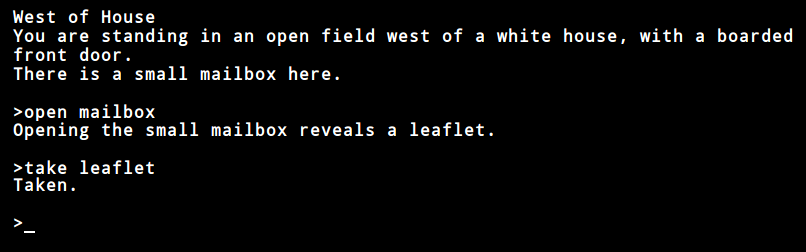}
        \caption{Intro to Zork}
        \label{fig:intro_zork}
    \end{figure}

    Consider Zork~\citep{if:Zork}, one of the genre's most famous examples. Figure~\ref{fig:intro_zork} depicts Zork's opening scene along with two player commands and the corresponding system responses.
    As illustrated, the game uses natural language to describe the state of the world, to accept actions from the player, and to report subsequent changes in the environment. Through sequential decision making, the player works toward goals which may or may not be specified explicitly.
    In the nomenclature of reinforcement learning (RL), language is the \emph{action space} and also the \emph{observation space}~\citep{narasimhan2015}.
    In text-based games, the observation and action spaces are both combinatorial and compositional -- major challenges for reinforcement learning.
    Furthermore, text-based games are partially observable since descriptive text does not communicate complete information about the underlying game state or may do so ambiguously.
    As a consequence of these (and other) challenges, hand-authored games like Zork are beyond the capabilities of current learning algorithms~\citep{narasimhan2015,haroush2018learning}.

    To help agents progress toward mastering text games in a controlled and scientific manner, we introduce the \emph{TextWorld} learning environment.
    TextWorld is a \emph{sandbox} environment~\citep{wright1996simcity,sukhbaatar2015mazebase} in which games of varying complexity emerge from a set of underlying world mechanics.
    In this setting, simpler games can act as stepping stones toward more complex games.
    Like the Arcade Learning Environment (ALE,~\citep{bellemare2013arcade}), Gym~\citep{brockman2016openai}, and CommAI~\citep{baroni2017commai},
    TextWorld enables interactive play-through of a curated set of games.
    Unlike previous text-based environments, including \href{https://github.com/danielricks/textplayer}{TextPlayer} and \href{https://github.com/MikulasZelinka/pyfiction}{PyFiction},
    TextWorld's sandbox functionality enables users to handcraft games or to construct games automatically through a suite of generative mechanisms.

    Specifically, TextWorld features a logic engine that automatically builds game worlds, populates them with objects and obstacles, and generates quests that define a goal state and how to reach it.
    It automatically generates text descriptions of underlying game states using an extensible vocabulary and a context-free grammar (CFG).
    Common-sense rules encoded in the logic and grammar govern generated worlds and the quests within them, to make these human-interpretable and consistent with existing games: \eg keys open locked doors and can be carried or stored in containers; food items can be combined, cooked, and eaten.
    Furthermore, because the vocabulary contains synonyms for most nouns, verbs, and adjectives, different surface forms can be applied automatically to abstract types to add variety and complexity: \eg the \entity{<container>} object may manifest as a \entity{chest} or \entity{cabinet}; the \cmd{<move>} action may manifest as \cmd{walk} or \cmd{go}.

    TextWorld's generative nature has several downstream implications for learning.
    First, it means there exists a known and structured representation of the partially observable game state.
    This enables exact state-tracking~\citep{henderson2014second} and the corresponding assignment of intermediate rewards in training (if desired).
    Second, agents can be trained on a potentially infinite set of related text games rather than a finite collection as in previous learning environments.
    By controlling parameters of the generative process for training and test games, TextWorld can be used to investigate curriculum learning, generalization, and transfer learning in RL.
    Tunable parameters include the length of quests, the size of environments, the number of abstract action and object types, the number of synonyms for each type, complexity of the descriptive grammar, and more.

    A powerful feature of language which motivates our interest in text games is that it abstracts away complex physical processes.
    For instance, through text an agent could learn and use the concept that \textsl{opening doors provides access to connected rooms} without going through the (literal) motions of turning knobs in 3D space and time.
    This level of abstraction can be useful for studying functions of control, planning, \etc~in isolation and in tandem with the function of language itself.

    The aim of this paper is to introduce TextWorld to the research community.
    Its primary contributions are:
    \begin{itemize}
        \item A survey of the machine-learning challenges of and approaches to text-based games, including a curated list of hand-authored games with corresponding analysis;
        \item A detailed description of the TextWorld framework, its features, and how to use it;
        \item An initial set of simple text games to be used as RL benchmarks;
        \item Evaluation of several baseline algorithms on both benchmark and hand-authored games.
    \end{itemize}
    Subsequent works will more deeply investigate novel approaches to RL for text games.
    Our hope is that TextWorld becomes a living resource, with contributors developing new benchmarks and algorithms to push the state of the art forward.

    The remainder of this paper is organized as follows.
    In Section~\ref{sect:rl} we introduce text-based games, formalize them as RL problems and highlight their challenges.
    In Section~\ref{sect:framework} we delve into details of the TextWorld framework, how it generates and interacts with text games, and how it may be used to train RL agents.
    Section~\ref{sect:related} describes related frameworks and existing approaches to solving text games, while Section~\ref{sect:benchmarks} describes some of TextWorld's benchmark tasks and our experimental results.
    We discuss limitations of the framework and future work in Section~\ref{sect:future} before concluding.

\section{Text Games from a Reinforcement Learning Perspective}
\label{sect:rl}

    Text-based games are sequential decision-making problems that can be described naturally by the Reinforcement Learning (RL) formalism. In this section, we define some of the terminology found in text-based games, formalize the text-based environment as an RL problem, discuss challenges faced by RL agents in such environments,
    and show how these challenges motivate the need for a framework like TextWorld.
    In the following, an ``agent'' is a model that takes text information as input and outputs text commands to progress through a game.

    \subsection{Text-based Games}
    \label{sect:text_based_games}
    Text-based games are turn-based games usually played through a command line terminal. At each turn, several lines of text describe the state of the game, and the player may enter a text command to change this state in some desirable way (\ie to move towards a goal). A game's built-in parser or interpreter deciphers player commands and maps them to state changes (events in the game). The genre became popular in the early 1980s especially with the release of Zork~\citep{if:Zork}, which featured rich storytelling and an advanced command parser.

      \subsubsection{Gameplay}
        \label{sect:parser_type}
\begin{figure}[ht]
          \centering
          \includegraphics[width=\textwidth]{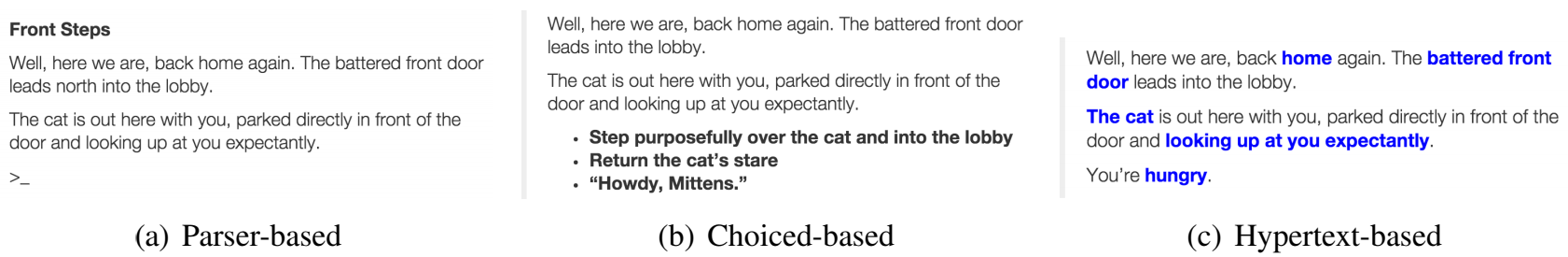}
          \caption{Types of text-based games. Image from~\citep{he2015deep}.}
          \label{fig:if_types}
        \end{figure}

        Text-based games can be classified according to how the player issues commands (see Figure~\ref{fig:if_types}): in \textbf{parser-based} games, the player types text commands character by character; in \textbf{choice-based} games, the player chooses from a given list of command options; and in \textbf{hypertext-based} games, the player clicks on one of several links present in the description. The work in this paper focuses on \textbf{parser-based} games.

        In text-based game terminology, each discrete in-game location is called a \emph{room} (\eg \literal{Kitchen}, \literal{In front of the white house}, and so on). A game may contain one or several rooms connected in some topology forming a \emph{map}. To explore the environment, the player issues navigational commands: \cmd{go} followed by a cardinal direction (\dir{north}, \dir{northeast}, \etc) or an orthogonal direction (\dir{up}, \dir{down}). 
        Maps vary greatly from one game to another. Room exits and entrances do not always match (\eg \cmd{go north} and then \cmd{go south} may not return you to where you started). Also, some navigational commands may not have a reciprocal (\eg in Zork~\citep{if:Zork}, a trapdoor locks from behind preventing the player from going back).

        Most of the time, rooms contain objects the player can interact with (\eg \cmd{take sandwich from table}, \cmd{eat sandwich}, \cmd{open door}, \etc). Objects are defined by their \emph{attributes}, which also determine how the player can interact with them (through object affordances). For instance, an object could be \predicate{portable} (\eg a \entity{lamp}) or \predicate{portable} and \predicate{edible} (\eg a \entity{sandwich}).

        One of the biggest challenges when playing parser-based games is figuring out what are the commands that will be understood by the parser (\ie that are intended by the game's author). Also, depending on the game, the result of some commands may be stochastic (\eg \cmd{go up} succeeds 75\% of the time, but 25\% of the time results in the player falling down to a previous room).
        A detailed list of puzzles and challenges traditionally found in text-based games can be found in Appendix~\ref{sect:appendix_puzzle_types_long}.

    \subsection{Text-based Games as POMDPs}
     \label{sect:rl:pomdp}
      Fundamentally, text-based games can be seen as partially observable Markov decision processes (POMDP)~\citep{kaelbling1998planning} where the environment state is never observed directly. To act optimally, an agent must keep track of all observations, \ie textual feedback received after entering commands. Although the observation can be augmented with feedback from commands like \cmd{look} and \cmd{inventory}, which describe the agent's surroundings and possessions, this information is still limited to the current room.

      Formally, a text-based game is a discrete-time POMDP defined by $(S, T, A, \Omega, O, R, \gamma)$, where $S$ is the set of environment states, $T$ is the set of conditional transition probabilities between states, $A$ is the set of words that are used to compose text commands, $\Omega$ is the set of observations, $O$ is a set of conditional observation probabilities, $R : S \times A \rightarrow \realdomain$ is the reward function, and $\gamma \in [0, 1]$ is the discount factor.

      \paragraph{Environment States ($S$)}
      The environment state at turn $t$ in the game is $s_t \in S$. It contains the complete internal information of the game, like the position and state of every entity (rooms, objects, player, \etc), much of which is hidden from the agent. When an agent issues a command $\bc_t$ (defined next), the environment transitions to state $s_{t+1}$ with probability $T(s_{t+1} | s_t, \bc_t)$.

      \paragraph{Actions ($A$)}
      At each turn $t$, the agent issues a text command $\bc_t$ of at least one word. In parser-based games, the interpreter can accept any sequence of characters (of any length) but will only recognize a tiny subset thereof. Furthermore, only a fraction of recognized commands will actually change the state of the world. The resulting action space is enormous and intractable for existing RL algorithms. We make the following two simplifying assumptions:
      \begin{itemize}
        \item \textbf{Word-level} Commands are sequences of at most $L$ words taken from a fixed vocabulary $\vocab$.
        \item \textbf{Syntax} Commands have the following structure: $\mathtt{verb [noun~phrase~[ adverb~phrase]]}$, where $\mathtt{[\dots]}$ indicates that the sub-string is optional. In this context, a $\mathtt{noun~phrase}$ is a string identifying an object (\eg \literal{the big wooden chest}). Similarly, an $\mathtt{adverb~phrase}$ provides additional context for the command (\eg \literal{with the red key}). To simplify the syntax further, determiners are omitted.\footnote{Typical text-based game interpreters disregard determiners.}
      \end{itemize}
      The agent's action space is some vocabulary $\vocab$ plus a special token \cmd{<return>} that indicates the end of a command. Each action $a^i_t \in A$ is a token, where $t$ is the turn in the game and $i$ indicates the $i$th token in the command $\bc_t$. A command is a sequence of $n \le L$ tokens $\bc_t = [a^1_t,\dots,a^n_t]$ that respects the syntax previously defined and ends with $a^n_t =$ \cmd{<return>}.

      The agent's policy is a mapping between its states and actions. In TextWorld, the agent's policy $\pi_{\theta}$, where $\theta$ are the policy's parameters, maps a state $s_t$ and words generated in the command so far to the next word to generate: $a^i_t = \pi_{\theta}(s_t,a^0_t,...,a^{i - 1}_t)$.

      \paragraph{Observations ($\Omega$)}
      The text information perceived by the agent at a given turn $t$ in the game is the agent's observation, $o_t \in \Omega$, which depends on the environment state and the previous command with probability $O(o_t|s_t,\bc_{t-1})$. In other words, the function $O$ selects from the environment state what information to show to the agent given the command entered. For instance, if the agent tries to open a chest, the observation returned by the environment might show that the chest is locked.


      \paragraph{Reward Function ($R$)}
      Based on its actions, the agent receives reward signals $r_t = R(s_t, a_t)$. The agent's goal is to maximize the expected discounted sum of rewards received $E \left[\sum_t \gamma^t r_t \right]$.

      Most text-based games (including those in our curated list) have a scoring mechanism whereby players receive points for completing (sub)quests and reaching new locations. When available, this score can be used as a reward signal. Otherwise, one could define reward signals by assigning a positive reward if the agent completes the game. Intermediate rewards might also be inferred from the interpreter's feedback. Note that this feedback usually only contains information about the results of individual commands (\eg \literal{I don't know this verb!}, \literal{This chest is locked!}) rather than about overall progress.

    \subsection{RL Challenges in Text-based Games}

    Complex environments make training RL agents challenging for several reasons. Here, we list some conventional challenges known in the RL literature that are also prevalent in text-based games.

      \paragraph{Partial Observability}
      As mentioned, states of text-based games are partially observable. Only the local information such as the current room description and the player's inventory is made available. Moreover, taking into account only the latest observation, it may be impossible to differentiate certain states based on observations. For instance, seeing a blue locked chest in a room, it is important for the agent to know whether or not it collected or saw a blue key in the past. The environment might give the same feedback for two different commands (\eg \literal{taken} might be the feedback for \cmd{take blue key} or \cmd{take red apple}). Furthermore, important information about the environment might not be apparent in the observation (\eg whether a chest is locked or not, what it contains, \etc). Observations may also be time-sensitive (\eg the agent only gets a reward when examining clues for the first time).

      \paragraph{Large State Space}
      With large state spaces, tabular methods for solving RL problems are no longer practical~\citep{sutton2018reinforcement}. Finding good approximate solutions is still an active area of research. In text-based games, the state space is combinatorial and enormous; the number of possible states increases exponentially with the number of rooms and objects.


      \paragraph{Large and Sparse Action Space}
      As with large state spaces, reasoning in an environment with a large number of actions necessitates finding good approximate solution methods to replace the tabular ones~\citep{sutton2018reinforcement}. The text-based game setting is especially challenging since the action space is large and sparse; the space of all word strings is much larger than the space of admissible commands (\ie commands that actually change the underlying state $s_t$). In addition, the outcome or even the validity of some commands might depend on a specific event or how much time has passed (\eg the tide rises and blocks the player in \citep{if:sherlock}).

      \paragraph{Exploration vs. Exploitation}
      Balancing exploration of the environment and the exploitation of known information is a fundamental issue in RL~\citep{mcfarlanesurvey}. Exploration is at the core of text-based games as they cannot be solved by learning a purely reactive controller. Instead, a strategy that promotes \emph{directed exploration} must be used; the agent must deliberately explore the environment, collecting information about objects and persons encountered along the way (\eg you never know what is in a box without opening it first). Such information hints about the goal/purpose of the game, what dangers are present, and provides clues that might become handy later in the game for solving puzzles. We expect that agents, like humans, will benefit from exploration driven by curiosity.

      \paragraph{Long-term Credit Assignment}
      Knowing which actions were responsible for obtaining a certain reward, especially when rewards are sparse, is another fundamental issue in RL~\citep{sutton2018reinforcement}. Sparse rewards are inherent to text-based games in which the agent must generate a sequence of actions before observing a change in the environment state or getting a reward signal. For instance, activating a switch might have no immediate effect although it is essential for completing the game. Most text-based games feature sparse rewards, on the order of a single positive reward every 10-20 steps when following an optimal state trajectory.





      \subsubsection{Additional Challenges}
        By their very nature, text-based games bring additional challenges related to natural language understanding.

        \paragraph{Observation Modality}
        Observations consist in the environment's textual feedback to the previous command. This means that the observation space is unbounded, consisting of arbitrary-length sequences of characters. To simplify things, we assume that an observation is made of space-separated words that may or may not be found in an English dictionary. One drawback of looking only at words is that we may lose some information provided by the spacing (\eg ASCII art in Infidel~\citep{if:infidel} or a sonar map in Seastalker~\citep{if:seastalker}).

        \paragraph{Understanding Parser Feedback}
        Text-based games process player input using a parser. The parser varies from game to game in terms of the actions it recognizes. For example, nearly all games recognize actions like \cmd{get}, \cmd{take} and \cmd{go}, but only some games recognize verbs like \cmd{tickle}, \cmd{swim}, \cmd{dig} and \cmd{bribe}. Part of the challenge of playing a parser-based text game is understanding which verbs and objects are recognized by the parser. Making this task more difficult, failure messages vary from game to game when the parser receives an invalid or unrecognized command.

        \paragraph{Common-sense Reasoning \& Affordance Extraction}
        To succeed at text-based games, it is necessary to understand how to interact with everyday objects. For example, if the description of a location includes a tree, it is likely that a player could climb or chop the tree, but not eat or drive it. The problem of identifying which verbs are applicable to a given object is called \emph{affordance extraction} and learning agents must solve it efficiently to make progress in text-based games without exhaustive search.

        \paragraph{Language Acquisition}
        Some objects and actions may be named with invented words. Also, modifier words affect how some objects can be interacted with (this is related to affordance extraction). The meaning of these words must be learned on-the-fly while interacting with the environment. Text-based games also use linguistic coreference, since it is more pleasant to humans, which can complicate the task for learning machines.

    \subsection{RL with TextWorld}
      Solving a single text-based game often corresponds to tackling most of the above challenges at once, which makes it very difficult for existing algorithms. What would be useful is a way of testing and debugging RL agents in simpler settings (\eg one room with two objects where the goal is to eat the edible one). This is the main purpose of TextWorld's generative functionality (described in Section~\ref{sect:generation}). It can be used to focus on desired subsets of the challenges listed above.

      First, it is possible to control the size of the state space (\eg the number of rooms, number of objects, and how many commands are required in order to reach the goal optimally). At the moment, TextWorld has deterministic transitions between states.

      It is also possible to control the partial observability of the state by augmenting the agent's observations. The environment can provide the agent with a list of objects present in-game or even provide all information about the current game state. For instance, instead of generating the observation that there is a blue chest, the environment could state that the chest is locked and that inside the chest there is a red apple. In this setting, the agent does not need to explore to determine the layout of the world and the objects it contains.

      TextWorld enables one to generate a large number of games and control their shared characteristics (map, objects, goals, \etc). This is useful for focusing, \eg on language acquisition and affordance extraction: the agent can be trained on games with a fixed number of rooms and object types but with different object names. By interacting with objects, the agent should learn which have a similar function and generalize from one game instance to another.

      There are several ways to ease the language generation task. It is possible to restrict the agent's vocabulary to in-game words only or to restrict the verbs that the agent can generate to those understood by the parser. It is also possible to use a simplified grammar where object names are replaced with symbolic tokens (\eg \literal{You see container1 and container2.}). Language generation can be circumvented completely by converting every generated game into a choice-based game. In this case, commands $c_t$ are the agent's actions, \ie the agent's output becomes an index into the set of admissible commands (see Section~\ref{sect:game_state}) rather a sequence of words.

      Finally, instead of earning rewards only at the end of a game if the agent is successful, one can also provide intermediate rewards during training based on environment state transitions and the ground truth winning policy (see Section~\ref{sect:intermediate_reward}).

\section{The TextWorld Learning Environment}
    \label{sect:framework}
    \begin{figure}
      \centering
      \includegraphics[width=0.8\textwidth]{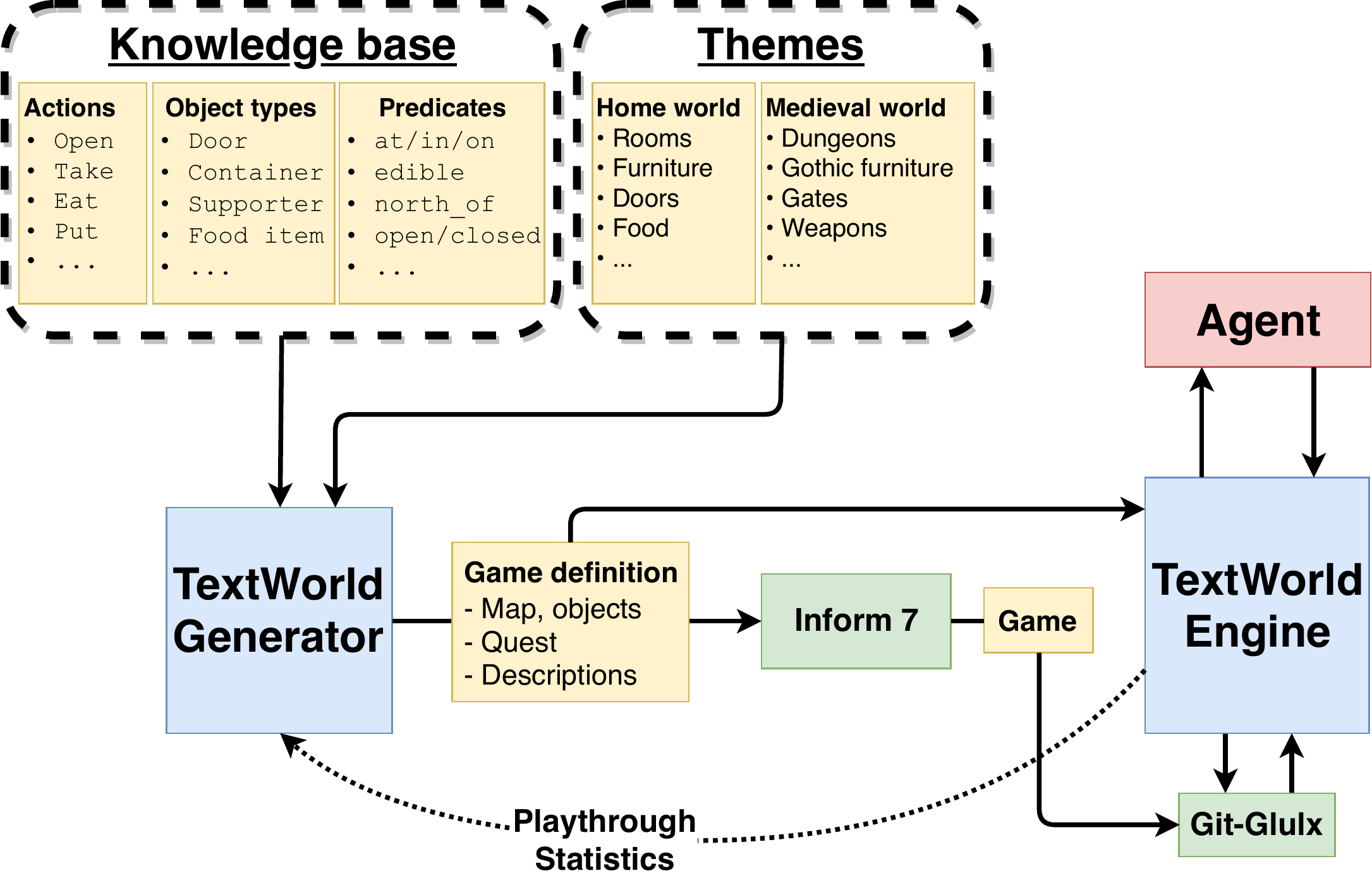}
      \caption[Architecture]{Overview of the framework. The two main components (in blue) of the proposed pipeline: the game generator and the game engine, which handles interactive play. Inform 7 and Git-Glulx are third-party libraries (in green) and the agent (in red) should be provided by the user. Given some knowledge base, sampled game definitions are first converted to Inform 7 code and compiled into a Glulx executable file. Then, agents interact with the game by communicating with the Git-Glulx interpreter via TextWorld.}
      \label{fig:architecture}
    \end{figure}

    TextWorld\footnote{Code and documentation can be found at \url{http://aka.ms/textworld}.} is a Python framework for training and testing RL agents on text-based games.
    It enables generation from a game distribution parameterized by the map size, the number of objects, quest length and complexity, richness of text descriptions, and more.
    The framework's architecture is shown in Figure~\ref{fig:architecture}. We detail the \emph{Engine} component in Section~\ref{sect:engine}, which covers how the internal state of generated games is represented and how the transition function is defined. The \emph{Generator} component is described in Section~\ref{sect:generation}, which explains the process of automatic game generation.
    TextWorld can also be used to play existing text-based games (see a curated list of games in Section~\ref{sect:curated_list}) but provides more limited information from the internal states of such games.

  \subsection{Game Engine}
  \label{sec:gameengine}
    \label{sect:engine}

    Game generation in TextWorld relies on a simple inference engine that ensures game validity at every step in the generation process. A game is said to be valid if it is possible to reach the end goal from the initial state. Although we could have used an existing problem-solver for this purpose, we did not require the full power of logical programming languages. TextWorld's inference engine implements simple algorithms specific to our environments, such as a one-step forward and backward chaining with or without fact creation (more on this in Section~\ref{sect:generation}). In future work our aim is to integrate the TextWorld framework with well established frameworks such as GDL~\citep{genesereth2005general} or STRIPS~\citep{fikes1971strips}.

\begin{figure}
        \centering
        \includegraphics[scale=0.4]{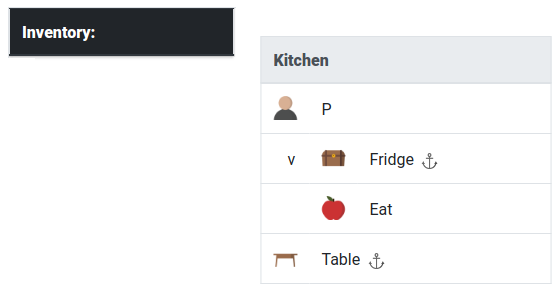}
        \caption{Simple environment with one room (\entity{kitchen}), a container (\entity{fridge}), a supporter (\entity{table}), a food item (\entity{apple}) and nothing in the player's inventory.}
        \label{fig:miniworld}
    \end{figure}

    To better explain the framework, let's consider the following simple text-based environment. There is a kitchen with a table and a closed fridge in which there is an apple. A visual representation can be seen in Figure~\ref{fig:miniworld}. The player is represented by the small avatar and the letter \entity{P}. Objects of the container type are represented by a chest, supporters (or surfaces) are represented by a table and food-type items are represented by an apple symbol. The anchor symbol next to certain objects means that they are fixed in place (\ie cannot be taken).

    TextWorld's generated text-based games can be represented internally as a Markov Decision Process (MDP)~\citep{Puterman:1994} defined by $(S, A, T, R, \gamma)$, where $S$ is the set of environment states, $A$ is the set of actions (with $A_{s_t}$ those available in state $s_t \in S$), $T(s_t, a, s_{t+1}) = \textrm{P}(s_{t+1}|s_t,a)$ is the state transition function that depends on current state $s_t$ and action taken $a \in A_{s_t}$, $R : S \times A \rightarrow \realdomain$ is the reward function, and $\gamma \in [0, 1]$ is the discount factor.
    Text-based games are \emph{episodic} since they stop when the player reaches one of the winning (goal) states $G \subset S$.

    Note that the MDP part of the POMDP, defined in Section~\ref{sect:rl:pomdp}, is semantically equivalent to the one described in this section. The sole exception is the action space; in the POMDP, we assume there exists an underlying function that maps text strings (generated by the agent) to the game actions defined in this section. This is the role of the game's interpreter.

    \paragraph{Environment states ($S$)}
    Game states are defined in terms of logical predicates. Each predicate \textbf{p}$(v_1, \dots, v_m)$ consists of a symbol \textbf{p} drawn from the alphabet of predicates $\Sigma$ followed by an $m$-tuple of variables. These \emph{predicates} define the relations between the entities (objects, player, room, \etc) present in the game. A \emph{logical atom} is a predicate whose variables are all are bound, \ie free variables (placeholders) have been substituted by concrete entities.

    A game state $s \in S$ consists in a multiset of logical atoms representing the \emph{facts} that are currently true about the world, also known as ``resources'' in linear logic. Taking as an example Figure~\ref{fig:miniworld}, the state depicted there can be represented as
    \begin{align*}
      s_t =~& \predicate{at}(\entity{fridge}, \entity{kitchen})
             \otimes \predicate{at}(\entity{table}, \entity{kitchen})
             \otimes \predicate{in}(\entity{apple}, \entity{fridge}) \\
            &\otimes \predicate{open}(\entity{fridge})
             \otimes \predicate{at}(\entity{P}, \entity{kitchen}),
    \end{align*}
    where the symbol $\otimes$ is the linear logic \emph{multiplicative conjunction} operator.

    The set of winning states $G$ is composed of any state $s$ for which all the winning conditions (a set of facts) hold. The winning conditions are determined during the game generation process (Section~\ref{sect:generation_quest}). For instance, a winning condition could be as simple as $\predicate{in}(\entity{apple}, \entity{I})$, \ie the apple being in the player's inventory. 

    \paragraph{State Transition Function ($T$)}
    The state transition function is defined using linear logic~\citep[Ch.~8]{russell2016artificial} and is inspired in part by Ceptre~\citep{martens2015ceptre}, a linear logic programming language. Logical rules are stored in a knowledge base and define what is possible in the game. For the working example, the relevant rules are
    \begin{align*}
      open(C)  &:: \$\predicate{at}(\entity{P}, R) \otimes \$\predicate{at}(C, R) \otimes \predicate{closed}(C) \multimap \predicate{open}(C)\\
      close(C) &:: \$\predicate{at}(\entity{P}, R) \otimes \$\predicate{at}(C, R) \otimes \predicate{open}(C) \multimap \predicate{closed}(C)\\
      take(F, C)   &:: \$\predicate{at}(\entity{P}, R) \otimes \$\predicate{at}(C, R) \otimes \$\predicate{open}(C) \otimes \predicate{in}(F, C) \multimap \predicate{in}(F, \entity{I})\\
      take(F, S)   &:: \$\predicate{at}(\entity{P}, R) \otimes \$\predicate{at}(S, R) \otimes \predicate{on}(F, S) \multimap \predicate{in}(F, \entity{I})\\
      put(F, S)    &:: \$\predicate{at}(\entity{P}, R) \otimes \$\predicate{at}(C, R) \otimes \$\predicate{open}(C) \otimes \predicate{in}(F, \entity{I}) \multimap \predicate{in}(F, C)\\
      insert(F, C) &:: \$\predicate{at}(\entity{P}, R) \otimes \$\predicate{at}(S, R) \otimes \predicate{in}(F, \entity{I}) \multimap \predicate{on}(F, S)\\
      eat(F, S)    &:: \predicate{in}(F, \entity{I}) \multimap \predicate{eaten}(F).
    \end{align*}

    The uppercase italic letters represent variables for objects of a certain type ($F$: food item, $S$: supporter, $C$: container and $R$: room). The entities \entity{P} and \entity{I} represent the player and its inventory. The symbol $\multimap$ (lolli) is the linear implication operator. The interpretation of the linear implication is such that it \emph{consumes} the resources on the \emph{left-hand-side} (henceforth LHS) and \emph{generates} resources on the \emph{right-hand-side} (henceforth RHS).
    The notation \$ is a shorthand meaning a predicate is implicitly carried over to the right-hand side. Given a state and the set of rules, we can perform \emph{forward chaining}. Applying a rule to state $s$, whose LHS is satisfied by $s$, leads to a conclusion, which is a new collection of atoms generated by the RHS of the selected rule.

    Applying all possible rules for all conclusions leads to a proof-tree with triplets $(s, a, s')$, where $s$ is an assumption and rule $a$ leads to the conclusion $s'$. Adding control to forward chaining, by only exploring paths where unseen states are introduced, leads to an algorithm that upon termination discovers all $s \in S$ in the MDP. Merging the duplicate states in the proof-tree provides the underlying $(S, A, T)$ of the MDP.


    \paragraph{Action Space ($A$)}
    An action is one of the rules defined in the knowledge base for which all free variables have been "grounded", \ie substituted by bound variables appropriately. Actions available in the current state, $A_{s_t}$, can be obtained by performing a single step of forward-chaining given facts true in $s_t$. In other words, the inference engine is used to retrieve all possible substitutions for every rule that can be applied to $s_t$. In the initial state of the working example, the available actions are
    \begin{align*}
      &close(\entity{fridge}) ::\\
      &~~\$\predicate{at}(\entity{P}, \entity{kitchen}) \otimes \$\predicate{at}(\entity{fridge}, \entity{kitchen}) \otimes \predicate{open}(\entity{fridge}) \multimap \predicate{closed}(\entity{fridge})\\
      &take(\entity{apple}, \entity{fridge}) ::\\
      &~~\$\predicate{at}(\entity{P}, \entity{kitchen}) \otimes \$\predicate{at}(\entity{fridge}, \entity{kitchen}) \\
      &~~\otimes \$\predicate{open}(\entity{fridge}) \otimes \predicate{in}(\entity{apple}, \entity{fridge}) \multimap \predicate{in}(\entity{apple}, \entity{I})\\
    \end{align*}

    \paragraph{Reward Function ($R$)}
      In the general case, games generated with TextWorld only provide a positive reward when reaching a winning state $s \in G$. The goal is to maximize the expected discounted sum of rewards received $E \left[\sum_t \gamma^t r_t \right]$ where $r_t = R(s_t, a_t)$.

    \subsubsection{Intermediate Reward}
      \label{sect:intermediate_reward}
      Tracking the state of the player allows us to determine a winning policy (not necessarily optimal) for any game state. A prerequisite is that a winning policy exists from the player's initial position (this is guaranteed for generated games). If so, then by monitoring state changes we can update the winning policy accordingly: if the agent performs the action dictated by the current winning policy, it progresses to the next desired state and we simply shift the policy forward one time-step; if the agent goes off the winning trajectory we add reciprocal actions to the policy to undo or correct this negative progress; and if the agent changes its state in a way that does not affect the quest, the winning policy does not change.

      In addition to the final reward, TextWorld can provide an intermediate reward which is tied to the winning policy. After each command, if the winning policy increases in length, meaning that as a result of the last action, additional commands are required to solve the game, then we assign a negative reward. If the winning policy shortens, meaning the last action brought the agent closer to the goal, we assign a positive reward. Otherwise, the reward is 0.

  \subsection{Game Generation}
    \label{sect:generation}
    TextWorld can be used as a sandbox environment in which predefined dynamics govern emergent games. With this sandbox, it is possible to generate a combinatorial (not to say infinite) set of games from which an agent could learn the underlying dynamics. Since we control the generation process, we can construct games where the knowledge to be learned is interpretable to humans.

    TextWorld's game generator takes as input a high-level specification of a game and outputs the corresponding executable game source code in the Inform~7 language (Appendix~\ref{sect:appendix_inform7}). The game specification assigns values to parameters such as the number of rooms, the number of objects, the length of the quest, the winning conditions, and options for the text generation (\eg theme, co-references, adjectives, and so on).

    \subsubsection{World Generation}
      \label{sect:generation_world}
      TextWorld generates maps through a simple procedure based on the Random Walk algorithm~\citep{pearson1905problem}. This enables us to generate a wide variety of room configurations, which in turn makes the set of possible games very large. The map generation process is parameterized by the number of rooms, the grid size of the world, and whether room connections should have doors or not. The grid size influences the compactness of the room configuration, where smaller grids mean more compact maps with potentially more loops.

      Once the map is generated, objects are added to the world uniformly across the rooms. Some objects are \predicate{portable} (as opposed to \predicate{fixed in place}), which means they can be nested on or in other objects that have the \predicate{supporter} or \predicate{container} attribute, or placed on the floor of a room or in the player's inventory.

    \subsubsection{Quest Generation}
    \begin{figure}
        \centering
        \begin{tabular}[t]{c|c|c}
            \includegraphics[width=0.3\textwidth,valign=T]{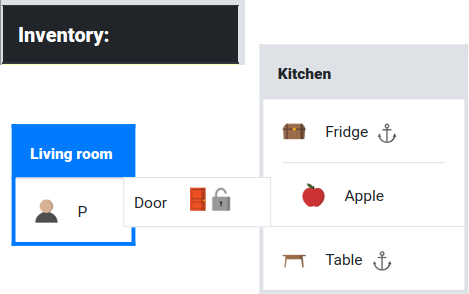} & \includegraphics[width=0.3\textwidth,valign=T]{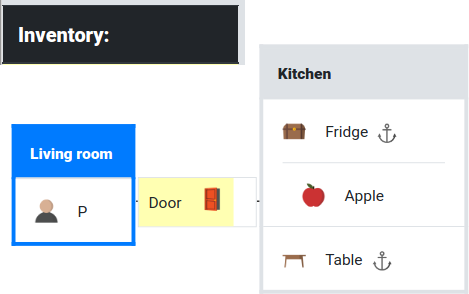} & \includegraphics[width=0.3\textwidth,valign=T]{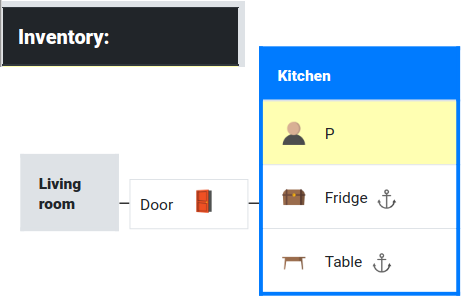}
            \\\hline
            \includegraphics[width=0.3\textwidth,valign=T]{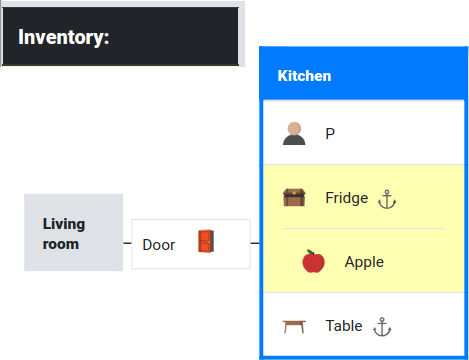} & \includegraphics[width=0.3\textwidth,valign=T]{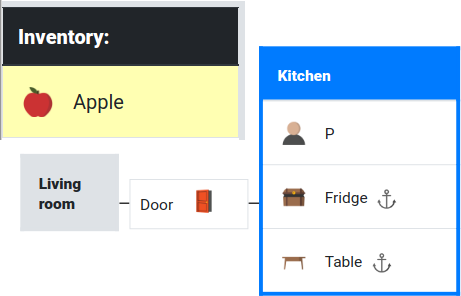} & \includegraphics[width=0.3\textwidth,valign=T]{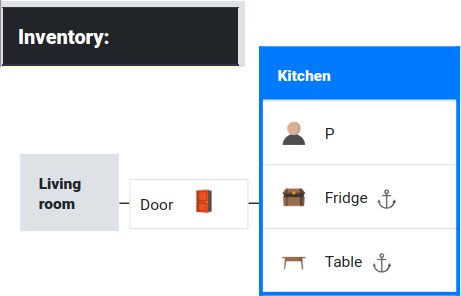}
        \end{tabular}
        \caption{Comic strip showing a simple quest where the player has to find and eat the apple.}
        \label{fig:comicstrip}
    \end{figure}
      \label{sect:generation_quest}
      In TextWorld, the purpose of any game is to reach a winning state. The term \emph{quest} will be used to represent a sequence of actions the player must perform to win the game. Note that this sequence does not have to be optimal or unique; many trajectories could lead to a winning state.

      We define \emph{quest generation} as the process of determining interesting sequences of actions from which to derive winning conditions. As discussed in Section~\ref{sect:engine}, the inference engine in TextWord can perform forward-chaining to construct a tree of all possible action sequences given an environment. However, not all paths are interesting (from a human or RL perspective). For this reason, we impose a dependency constraint on the actions and reject paths containing cycles. The dependency relation is defined as follows: action $a_t$ depends on action $a_{t-1}$ if and only if the RHS of $a_{t-1}$ generates the resource(s) required by the LHS of $a_t$. The winning condition of a given quest is the set of resources generated by the RHS of the last action.

      We can generate quests by modifying the forward chaining algorithm to apply these constraints, calling the resulting process \emph{forward quest generation}. An example quest is depicted in Figure~\ref{fig:comicstrip}. First, the player opens the door and moves south to enter the kitchen. When in the kitchen, the player opens the fridge, takes the apple, and finally eats it.

      \paragraph{Backward Quest Generation}
      The end goal often defines the nature of a quest and yields significant rewards for the agent. Thus, it is desirable to specify the end goal in quest generation.
      The forward quest generation algorithm indirectly allows this specification, by generating all possible quests from an initial condition then searching for those that satisfy the ending constraint. However, as the number of states and the length of the desired quest increases, this approach becomes intractable. To remedy this, TextWorld also supports backward chaining. Backward chaining simply reverses forward chaining, starting from a specified end state rather than an initial state. The same dependency between subsequent actions and cycle rejection apply.

        \begin{figure}
            \centering
            \begin{tabular}{c|c|c|c}
                \includegraphics[width=0.24\textwidth]{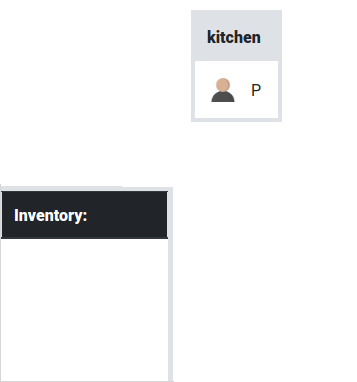} & \includegraphics[width=0.23\textwidth]{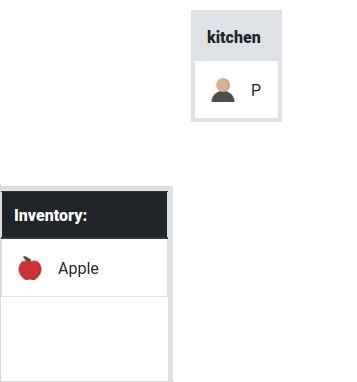} &
                \includegraphics[width=0.24\textwidth]{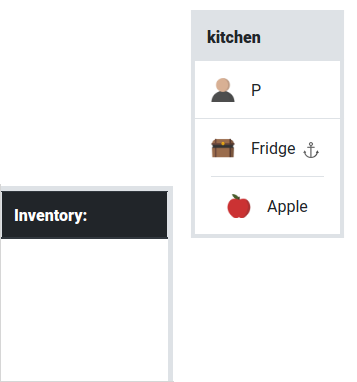} & \includegraphics[width=0.23\textwidth]{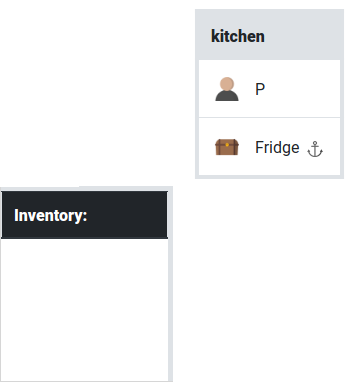}
            \end{tabular}
            \caption{Comic strip showing backward quest generation for the quest \cmd{open fridge} / \cmd{take apple from fridge} / \cmd{eat apple}. The player is first placed in the kitchen. The apple is created and placed in the player's inventory at Step 2. The fridge is created at Step 3, and the apple placed within it. In the last step the fridge is closed. This becomes the game's starting state.}
            \label{fig:comicstrip_quest_generation}
        \end{figure}

      \paragraph{Extending the World during Quest Generation}
      TextWorld's generator extends forward and backward chaining with a \emph{fact creation} step, which may occur before sampling a subsequent (or previous) action. Through fact creation, the generative process can add missing objects and facts to the world as needed, which yields more diverse quests and games. Figure~\ref{fig:comicstrip_quest_generation} shows a simple example of backward quest generation.

    \subsubsection{Text Generation}
      \label{sect:generation_text}
      The Text Generation module takes logical elements of the game state and renders them into coherent text. This serves as the observation provided to the agent. The engine generates object names, room descriptions, and quest instructions in constrained natural language using a context-free grammar (CFG)~\citep{Chomsky_1956}.
      The ease of authoring such grammars has led to their adoption in various natural language generation (NLG) systems for games~\citep{ryanExpressionist2016}.

      The module is essentially a set of grammars, each generating particular aspects: \eg there are separate grammars for creating object names, room descriptions, and instructions, respectively.
      Observable text is generated by iterating over all elements of the observation and filling templates with the appropriate information.
      As an example for object description:
      For a red box, the grammar may return \literal{There is a [object-noun] here. It is [object-adjective]}., which is filled to create \literal{There is a box here. It is red.}
      Some basic maintenance also ensures fluency, \eg using ``an'' vs. ``a'' when a noun begins with a vowel.

     Using a context-free grammar gives a degree of textual variation, allowing the same world and quest to be represented a number of ways while also ensuring strict control over the results.
     While our current grammars function like a text templating system, CFGs offer the possibility of deeper, recursive text generation should it be desired.
     Our grammars can be extended and modified easily with additional production rules, enabling the generation of simpler or more complex sentence structures that may act as a level of game difficulty.


      \paragraph{Object Names} Object names are assigned to each term in the game. Names are randomly picked from the CFG and uniquely assigned to objects. The object's type is used to derive the start symbol sent to query the CFG for a name. An object name can be decomposed into two parts: adjective and noun.
      The adjective is optional, and may be used to create more complex object names, and correspondingly more complex descriptions. Object name generation is in general straightforward, and consists of selecting a random adjective and noun and combining them. So, \eg given the nouns \literal{box} and \literal{cup}, as well as the adjectives \literal{dusty} and \literal{red}, there are four possible object names, \literal{dusty box}, \literal{red box}, \literal{dusty cup} and \literal{red cup}. Adjectives are also used as hints to the player; for example, a key's adjective will always match the adjective of what it opens, \eg a \literal{red key} will open the \literal{red chest}.


      \paragraph{Room Descriptions} The description of a room is the concatenation of the \emph{room-level} description of every object it contains, shown typically when entering the room or upon using the \cmd{look} command. The room-level description of an object contains information the player should be aware of upon entering the room (\eg \literal{There is a chest here. It is open and you can see some gold coins in it.}). The room's description also mentions its possible exits (\eg \literal{There is a path leading north.}). It is updated dynamically based on changes to the states of objects in the room, for example listing whether a container is open, closed, or locked, and which objects it contains.

      \paragraph{Quest Instructions} We use instructions to explain to the player what to do in a game. An instruction is a piece of text describing a particular action or several different actions. For example, \literal{Retrieve the blue key} could be used to represent the action \cmd{take blue key}, whereas \literal{Take the red key from the locked chest} may represent the sequence of actions \cmd{unlock chest} / \cmd{open chest} / \cmd{take red key}. In TextWorld, instructions may optionally describe every action of a quest (easier), only the final action (harder), or they may force the player to figure out what to do from scratch (goal identification; hardest). Likewise, the ability to combine actions into a single instruction can also be toggled; identifying a sequence of actions from an instruction rather than a single action is an additional challenge.

      \paragraph{Text Generation Options}
      TextWorld offers some control over different aspects of the text generation. Objects with similar attributes/states can be grouped together when describing a room (\eg \literal{In here, you see two red containers: a box and a chest.}). Objects mentioned in an instruction can be referred to using one or several of their attributes (\eg \literal{Take the red edible thing.}). Use of coreference (\eg \literal{There is a chest. It is open. In it, you see nothing interesting.}) is also optional.

      \begin{figure}
          \centering
          \subfigure[House]{
            \includegraphics[width=0.48\textwidth]{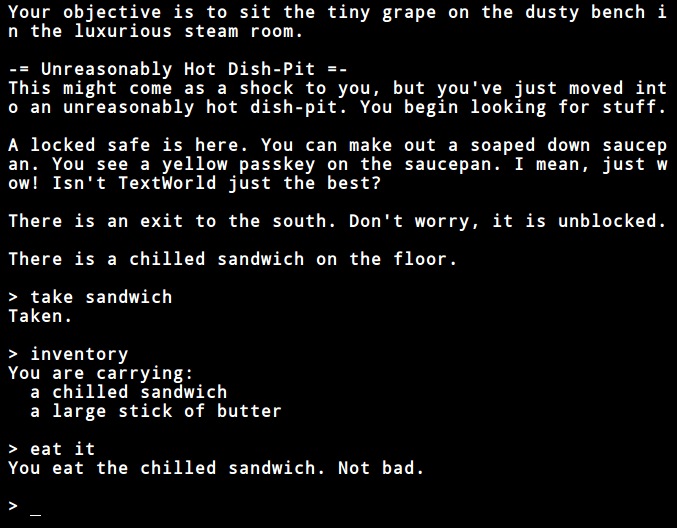}
          }
          \subfigure[Basic]{
            \includegraphics[width=0.48\textwidth]{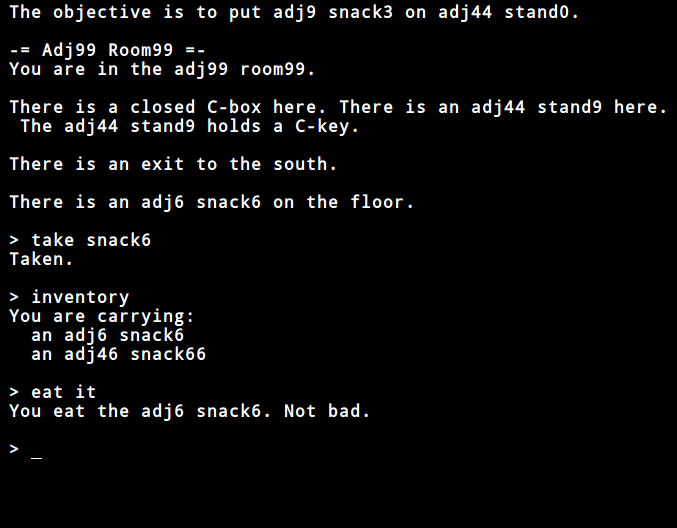}
          }
          \caption[Basic vs. house]{
            The same generated game with two themed grammars: house and basic.
          }
          \label{fig:games_with_different_themes}
      \end{figure}

      TextWorld also offers the choice between two themed grammars: house and basic. The house theme describes the world as if the game takes place in a modern house. The second theme uses a simple grammar with almost no linguistic variation (\eg no adjectives, no multi-word names). In this case, objects with the same attributes use a shared, prototypical prefix for their names followed by a number (\eg \entity{stand42}). The basic grammar cuts down the vocabulary and the language complexity to ease the training of neural generative models. These house and basic themes can be seen applied to the same underlying game in Figure~\ref{fig:games_with_different_themes}.

      \begin{figure}
        \centering
        \includegraphics[width=0.8\textwidth]{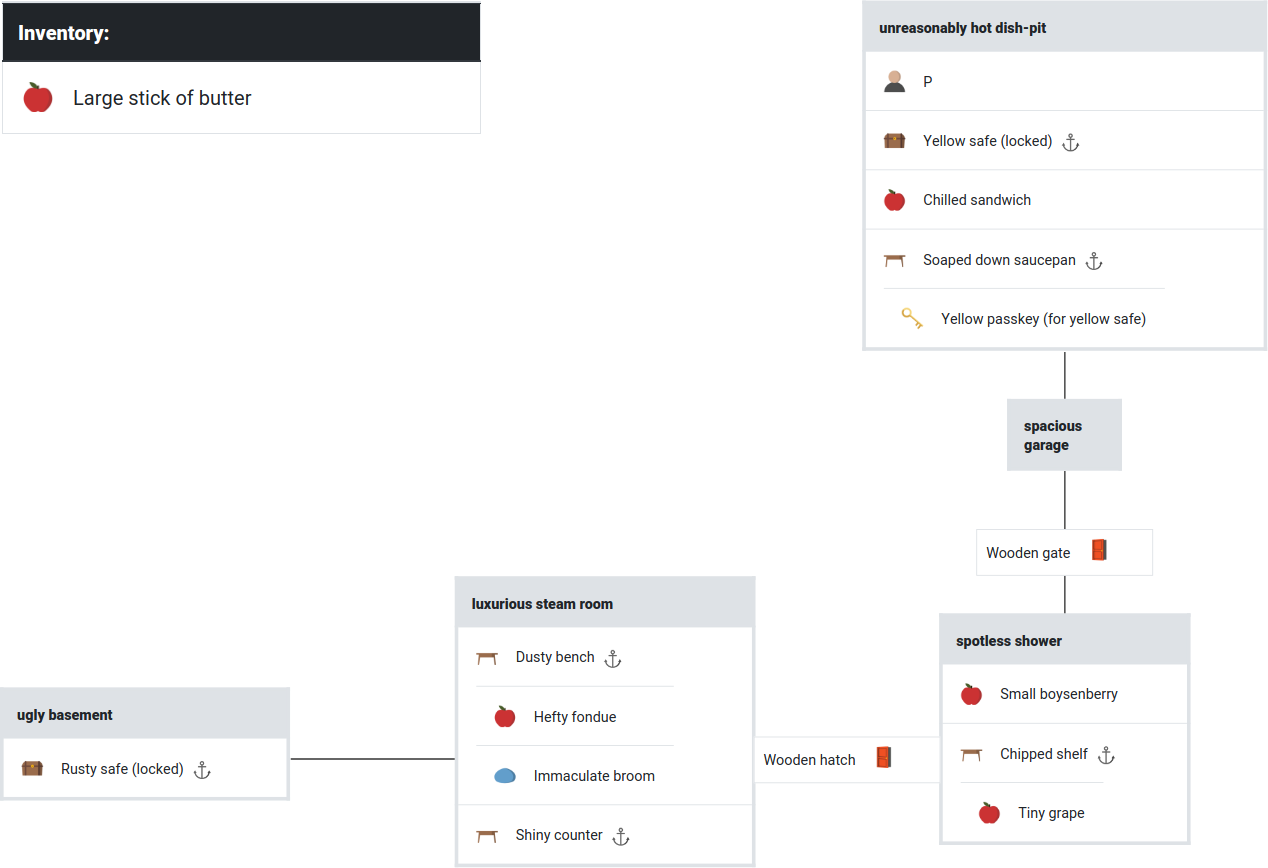}
          \caption[Visualization]{
            Visualization of the game shown in Figure~\ref{fig:games_with_different_themes} using TextWorld's integrated game-state viewer. The winning sequence of commands for this game: \cmd{go south} / \cmd{go south} / \cmd{take tiny grape from chipped shelf} / \cmd{go west} / \cmd{put tiny grape on dusty bench}.
          }
          \label{fig:visualization}
      \end{figure}

  \subsection{Game Interaction with TextWorld}

    Most basically, TextWorld can be used to play any text-based games that are interpretable either by Z-machine (via the Frotz interpreter) or by Glulx (via a custom git-glulx interpreter). The framework handles launching and interacting with the necessary game processes. It provides a simple API for game interaction inspired by that of OpenAI's Gym~\citep{brockman2016openai}. As an example, playing a text-based game requires just a few lines of Python code:
    \begin{minted}{python}
    import textworld
    env = textworld.start("zork1.z5")
    game_state = env.reset()  # Reset/initialize the game.
    reward, done = 0, False
    while not done:
      # Ask the agent for a command.
      command = agent.act(game_state, reward, done)
      # Send the command to the game and get the new state.
      game_state, reward, done = env.step(command)
    \end{minted}


    \subsubsection{Game Observation}
        \label{sect:game_state}
        The \code{game\_state} object contains useful information\footnote{See the framework's documentation for a list of all information obtainable from the game state object.} such as:
        \begin{description}
          \item[Feedback] The interpreter's response to the previous command, \ie any text displayed on screen.
          \item[Description] The description of the current room, \ie the output of the \cmd{look} command.
          \item[Inventory] The player's inventory, \ie the output of the \cmd{inventory} command.
          \item[Location] The name of the current room.
          \item[Score] The current score.
        \end{description}

        Game state information is fairly limited when it comes to existing games, whereas games generated with TextWorld can provide much more information if desired. Additional information includes:
        \begin{description}
          \item[Objective] Text describing what the player must do to win the game.
          \item[Admissible Commands] A list of commands that are guaranteed (i) to be understood by the interpreter and (ii) to affect the game state or to return information of the game state. Using this information in any way corresponds to playing a \emph{choice-based} game rather than a parser-based one.
          \item[Intermediate Reward] A numerical value representing how useful the last command was for solving the game (as described in Section~\ref{sect:intermediate_reward}).
          \item[Winning Policy] A list of commands that guarantees winning the game starting from the current game state.
        \end{description}


\section{Related Work}
  \label{sect:related}
    Text-based games are hard to solve, even for humans, because they require language understanding, planning, and efficient exploration to a greater degree than perception and reflexes (like most Atari games). Nonetheless, a few researchers have tried different approaches that we report in Section~\ref{sect:related_models}. Since TextWorld is a new learning environment, we compare it to relevant frameworks in Section~\ref{sect:related_frameworks}.

    \subsection{Relevant Models}
      \label{sect:related_models}
      \cite{narasimhan2015} develop a two-stage model called LSTM-DQN for parser-based text games, using the deep Q-learning framework. Their model encodes the agent's observation $o$ (a sequence of words) using mean-pooled LSTM states. The resulting state representation is used by two sub-networks to predict the Q-value over all verbs $w_v$ and object words $w_o$ independently. The average of the 2 resulting scores gives the Q-values for the composed actions.

      \vspace{-2em}
      \begin{align}
        Q(s, (w_v, w_o)) = \frac{Q(s, w_o) + Q(s, w_v)}{2}  \\
        Q(s, w_o) = MLP_o(s), \; Q(s, w_v) = MLP_v(s) \\
        s = LSTM(o), \; \text{where}~o = w_1 \ldots w_n
      \end{align}
      \vspace{-2em}

      They test their approach on 2 environments -- Homeworld and Fantasyworld -- using the Evennia toolkit\footnote{\url{http://www.evennia.com/}}. Homeworld is a small environment with 4 rooms, a vocabulary of 84 words, and 4 quests. Quests are also specified through text; for example, \literal{Not you are sleepy now but you are hungry now} (which indicates that the player should obtain food but should not get into bed).
      Fantasyworld is much larger, with a vocabulary size of 1340, and has stochastic state transitions. The LSTM-DQN completes 100\% of the quests in Homeworld and 96\% of quests in Fantasyworld. The authors also perform transfer-learning experiments by defining Homeworld2, which is made by shuffling the rooms and paths from Homeworld. The LSTM-DQN is trained on Homeworld and the LSTM component is transferred to Homeworld2. The transferred agent learns faster than one without training on Homeworld.

      \cite{he2015deep} introduce the Deep Reinforcement Relevance Network (DRRN) for tackling choice-based text games. They evaluate the DRRN on a deterministic game called ``Saving John'' and a larger-scale stochastic game called ``Machine of Death''. These games have vocabulary sizes 1762 and 2258 and action vocabulary sizes of 171 and 419, respectively.
      The DRRN takes as input the observation $o$ of the state $s$ and action choices $a_j$ and computes a Q-value for each possible pair:

      \vspace{-2em}
      \begin{align}
        Q(s, a_j) = g(f_s(o) , f_a(a_j)), \; \text{where}~o = w_1 \ldots w_n
      \end{align}
      \vspace{-2em}

      The DRRN model converges on both games when trained with the DQN algorithm with experience replay and Boltzmann exploration. It achieves optimal cumulative-reward on Saving John and a suboptimal but stable policy on Machine of Death. The authors test the DRRN trained on Machine of Death on state-action pairs paraphrased by participants. They show high correlation between the original and paraphrase $Q(s, a, \theta)$.

      Note that neither the LSTM-DQN nor the DRRN conditions on previous actions or observations. This means that neither has the capacity to deal with partial observability.

      Related work has been done to reduce the action space for parser-based games. \cite{haroush2018learning} introduce the Action Elimination Network to estimate the probability of an action failing in a given scene. To achieve this, feedback from the game engine is also stored in the replay buffer in addition to the standard \textit{<observation, action, reward>} triplets.
      The elimination module is trained with the stored quadruplets and assigns a score to each element in the large set of actions. During $\epsilon$-greedy exploration, at the greedy step the agent is only allowed to consider the top-$k$ actions, while during exploration, random actions are rejected with a predefined probability if their score is below a threshold.

      \cite{fulda2017can} tried to accomplish something similar by training word embeddings to be aware of verb-noun affordances. From that embedding, they manually select a group of verb-noun pairs for which they assume the vector $emb(noun) - emb(verb)$ encodes the affordance relation. Using the average of such vectors gives them an ``affordance vector'' that can be used to project the embedding of new nouns to a region of the embedding space where relevant verbs should be found.

      \cite{kostka2017text} build an agent specifically targeting the domain of classic text-based games. They pre-train an LSTM language model on fantasy books that
      is used in their model to extract important keywords
      from scene descriptions. Furthermore, they collect a
      list of possible commands from game solutions and
      semi-automatically extract a large number of commands from online tutorials and decompiled game sources.
      Their system
      follows a modular design where each typical phase of text-adventure gameplay is modeled by a separate algorithm:
      command generation, battle mode, inventory management, exploration, restart.
      Their model generates commands by
      finding keywords in the scene text and cross-referencing the extracted command corpus to find plausible commands.


    \subsection{Frameworks}
      \label{sect:related_frameworks}
      In the fields of AI in general and RL in particular, games have played a major role in advancing the state of the art.
      The well-known Arcade Learning Environment (ALE)~\citep{bellemare2013arcade},
      which provides an interface to Atari 2600 games,
      led to human-level videogame play by deep RL algorithms~\citep{mnih2015human}.
      One limitation of ALE, however, is that it does not facilitate research in generalization, metalearning, and transfer learning because the individual games are too distinct from one other. Indeed, most research using ALE focuses on training a separate agent (with the same architecture) for each game~\citep{machado2017revisiting}.

      ALE-style game collections exist in contrast to \emph{sandbox} environments,
      in which games emerge from a set of underlying world mechanics.
      Perhaps the best known such environment is SimCity~\citep{wright1996simcity}.
      Sandboxes generate a series of games that share components, and thereby naturally overcome some limitations of collections. Sandbox environments also allow for the programmatic construction of game variants whose difficulty can be tuned to form a curriculum.

      The MazeBase environment~\citep{sukhbaatar2015mazebase}, possibly the sandbox framework most similar to TextWorld, enables researchers to generate two-dimensional grid-based games.
      Each grid point may contain a certain object type, such as an obstacle, a pushable block, a switch, or the goal.
      Maps  are provided in an egocentric text-based representation to the player.
      One of the main motivations of MazeBase is to foster research in learning algorithms that reuse knowledge and policies acquired in related environments and games.
      Quests and the language in TextWorld are significantly more complex than in MazeBase.

      The CommAI framework~\citep{baroni2017commai} emphasizes the ability to generate curricula~\citep{bengio2009curriculum}, so that agents may learn incrementally from environments of increasing complexity.
      Interaction with the environment takes place at the lower level of bits rather than
      simplified natural language as in TextWorld.


      Recently, multimodal visuo-linguistic environments were introduced to study grounded language learning through RL.~\cite{chaplot2017gated} customized the VizDoom platform~\citep{kempka2016vizdoom} for language grounding experiments: objects with certain properties are placed in a room and the agent is instrcuted which object(s) to find. To perform similar experiments,~\cite{hermann2017grounded} add a language instruction component to the DeepMind Lab 3D navigation environment~\citep{beattie2016deepmind}.

\section{Benchmarks}
  \label{sect:benchmarks}
    In this section, we describe benchmark games that can be used through TextWorld to evaluate RL agents and algorithms. This set is preliminary; we plan to develop increasingly complex benchmarks in future work.

    \subsection{Curated List}
      \label{sect:curated_list}
      Following \cite{fulda2017can}, we compiled a list of 50 hand-authored text games to use as an evaluation set. Games designed for human players require, and can be used to measure, general capabilities like common-sense reasoning, planning, and language understanding. We manually analyzed all games in the set to ensure they are valid, with scores and interesting quests. From the original 50 proposed by \cite{fulda2017can}, we replaced 20 which were either parodies of text-based games or did not provide scores. The information we collected for the games can be found in Appendix~\ref{sect:appendix_curated_list}.

        \begin{figure}
            \centering
            \includegraphics[width=0.8\textwidth]{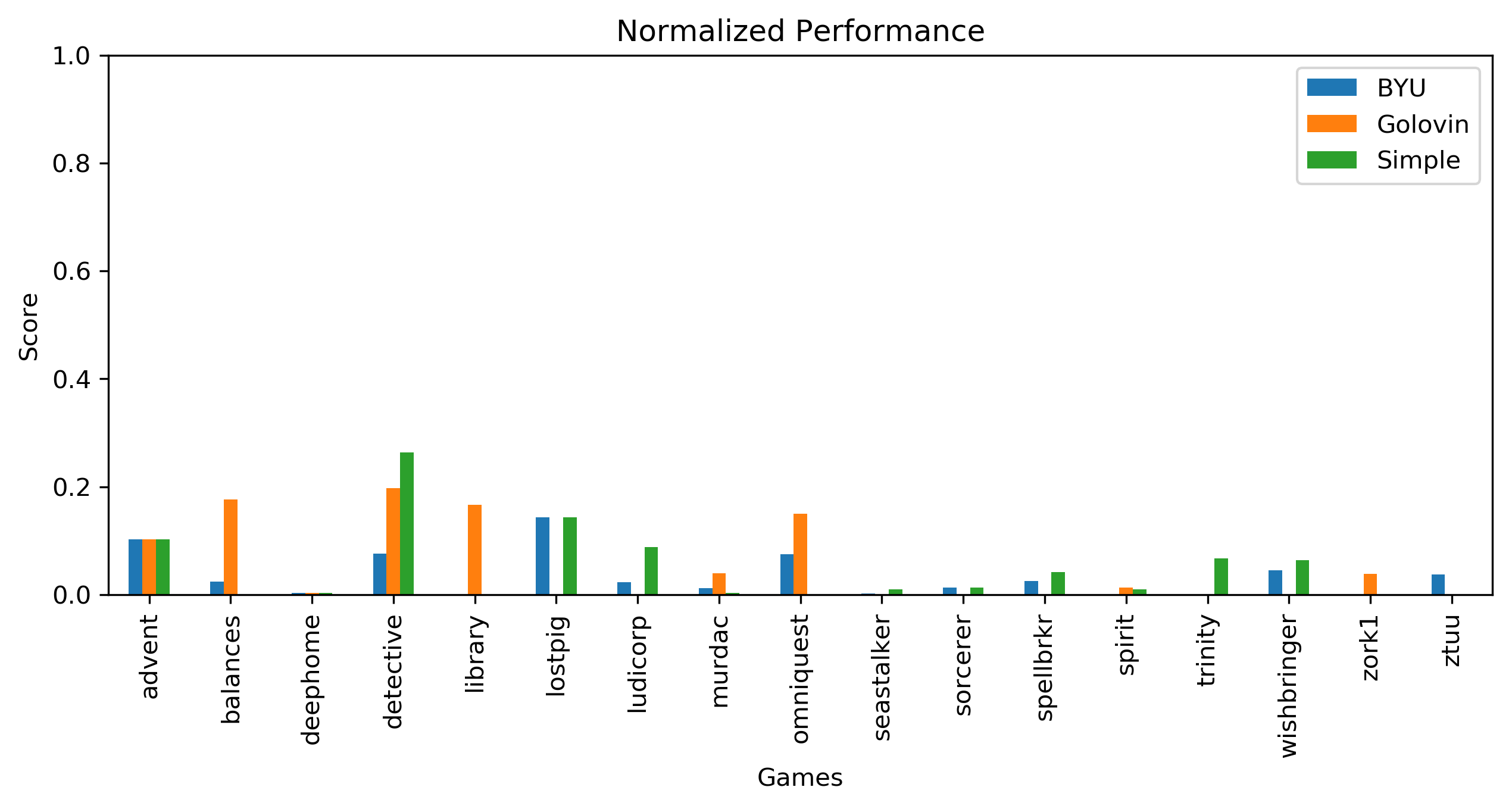}
            \caption{Normalized score for baselines evaluated on the curated list. Games where no score was obtained are omitted.}
            \label{fig:curated_list_results}
        \end{figure}

        We evaluate three baselines on the proposed list. BYU\footnote{Implementation from \url{https://github.com/danielricks/BYU-Agent-2016}.} and Golovin\footnote{Implementation from \url{https://github.com/Kostero/text_rpg_ai}.} are agents developed by \cite{fulda2017can} and \cite{kostka2017text}, respectively, and are described in Section~\ref{sect:related_models}. Both models were submitted to the Text-Based Adventure AI Competition~\citep{atkinson2018text} which consisted in playing 20 hidden games. The third is the \emph{Simple} baseline, which consists in sampling uniformly a command from a predefined set\footnote{The exact set is: \cmd{north}, \cmd{south}, \cmd{east}, \cmd{west}, \cmd{up}, \cmd{down}, \cmd{look}, \cmd{inventory}, \cmd{take all}, \cmd{drop} and \cmd{YES}.} at every step.

        Each agent has 1000 steps to get a high score. If the agent loses the game, the game is reset and play resumes until the step limit is reached. If the agent wins (which never happens in practice), the game stops and the evaluation is done. The procedure proposed here is slightly different from that in \cite{fulda2017can}, where they allow agents to play each game 1000 times for a maximum of 1000 steps each (\ie a theoretical limit of a million interactions). Our main motivation for having a small ``time'' budget is that we are interested in measuring the generalization capabilities of agents. When using this evaluation, we assume the agents are already trained (on other similar games) or encompass some prior knowledge (which is the case for the three baselines).

        Figure~\ref{fig:curated_list_results} reports the normalized score (\ie maximum score achieved divided by the game's max possible score) for each baseline agent on the curated list. Unsurprisingly, agents achieve a rather low score on a few games and zero on many. Using the information we gathered during our analysis, we can make educated guesses as to why some baselines perform well on certain games. For instance, in Advent the player starts with 36 points, which explains why all three baselines have the same score. As another example, the Detective game can be solved with mostly navigational commands. This explains why the Simple agent performs relatively well, since the commands it samples from are mostly navigational.

    \subsection{Treasure Hunter}
      \label{sect:treasure_hunter}

      This benchmark is inspired by the task proposed in \cite{parisotto2017neural}, where the agent spawns in a randomly generated maze and must find a specific object.
      A colored ``indicator'' object near the agent's starting position determines which object the agent must retrieve.
      The agent earns a positive reward for retrieving the correct object or a negative reward for an incorrect object. There is a limited number of turns.

      We adapted this task, which takes place in a 3D environment, to TextWorld.
      In our setting, the maze is a randomly generated map (see Section~\ref{sect:generation_world}) of rooms.
      We randomly place the agent and two objects on the map.
      Then, we randomly select which object the agent should recover and mention it in the welcome message (our indicator).
      In navigating to and obtaining the desired object, the agent may have to complete other tasks like finding keys and unlocking doors.

      The aim of this benchmark is to assess skills of affordance extraction (agents should determine verb-noun pairs that change the environment state);
      efficient navigation (agents should avoid revisiting irrelevant rooms); and
      memory(agents should remember which object to retrieve).

      We define the difficulty levels for this benchmark as follows:
      \vspace{-1em}
        \begin{itemize}
          \item \textbf{ 1 to 10:} Mode: easy,   \#rooms =  5, quest length linearly increasing from 1 to  5;
          \item \textbf{11 to 20:} Mode: medium, \#rooms = 10, quest length linearly increasing from 2 to 10;
          \item \textbf{21 to 30:} Mode: hard,   \#rooms = 20, quest length linearly increasing from 3 to 20;
        \end{itemize}
      where the modes are defined as
      \vspace{-1em}
        \begin{itemize}
          \item \textbf{Easy:} Rooms are all empty except where the two objects are placed. Also, connections between rooms have no door;
          \item \textbf{Medium:} Rooms may be connected by closed doors. Container objects are added, and might need to be opened to find the object;
          \item \textbf{Hard:} Locked doors and containers are added which may need to be unlocked (and opened) to reach the object.
        \end{itemize}

        Note that beyond the predefined difficulty levels, this benchmark can be simplified by letting the agent directly tap into the game state information (\eg feedback, description, inventory and objective) or using a simpler grammar.

        \subsubsection{Evaluation: One-life Game}
            One of our desiderata in building TextWorld is the ability to generate unseen games, so that we can train and evaluate an agent's performance on different game sets. To support our claim of this need, we test two state-of-the-art agents on a set of games generated by TextWorld, where the agents only see each game once and therefore cannot memorize them.

            Specifically, for each difficulty level described above, we generate 100 games. We run each agent on these games for a maximum of 1000 steps. For each game, when the agent picks up either the right object or the wrong one, it receives +1 or -1 score, respectively, and the game terminates immediately; if the agent exhausts all 1000 steps without finding any object, it receives 0 score.

            Evaluation results are provided in Table~\ref{tab:th}, where we compare a choice-based\footnote{Because of the compositional properties of language, a random parser-based agent would perform poorly since most generated commands would not make sense. In this work we use a choice-based random agent. It is not directly comparable to the other two agents, but it can give a general idea how difficult the games are in different difficulty levels.} random agent (\ie at each game-step the baseline model randomly selects one command from the list of admissible commands), the BYU agent and the Golovin agent. We report the average score for different difficulty levels and the average number of steps it took to finish the games (either win, lose, or exhaust).


            \begin{table}[t!]
            \small
              \resizebox{1.0\textwidth}{!}{
                \begin{tabular}{r|cc|cc|cc}
                  \toprule
                  & \multicolumn{2}{c|}{Random} & \multicolumn{2}{c|}{BYU} & \multicolumn{2}{c}{Golovin}\\
                  Model & Avg. Score & Avg. Steps & Avg. Score & Avg. Steps & Avg. Score & Avg. Steps\\
                  \midrule
                  \texttt{level  1} & 0.35 & 9.85 & 0.75 & 85.18 & \textbf{0.78} & 18.16 \\
                  \texttt{level  5} & \textbf{-0.16} & 19.43 & -0.33 & 988.72 & -0.35 & 135.67 \\
                  \texttt{level 10} & -0.14 & 20.74 & \textbf{-0.04} & 1000 & -0.05 & 609.16 \\
                  \texttt{level 11} & \textbf{0.30} & 43.75 & 0.02 & 992.10 & 0.04 & 830.45 \\
                  \texttt{level 15} & \textbf{0.27} & 63.78 & 0.01 & 998 & 0.03 & 874.32 \\
                  \texttt{level 20} & \textbf{0.21} & 74.80 & 0.02 & 962.27 & 0.04 & 907.67 \\
                  \texttt{level 21} & \textbf{0.39} & 91.15 & 0.04 & 952.78 & 0.09 & 928.83 \\
                  \texttt{level 25} & \textbf{0.26} & 101.67 & 0.00 & 974.14 & 0.04 & 931.57 \\
                  \texttt{level 30} & \textbf{0.26} & 108.38 & 0.04 & 927.37 & 0.04 & 918.88 \\
                  \bottomrule
                \end{tabular}
              }
              \caption{Model performance on one-life treasure hunter tasks.}
              \label{tab:th}
            \end{table}


        \begin{figure}
            \centering
            \includegraphics[width=0.8\textwidth]{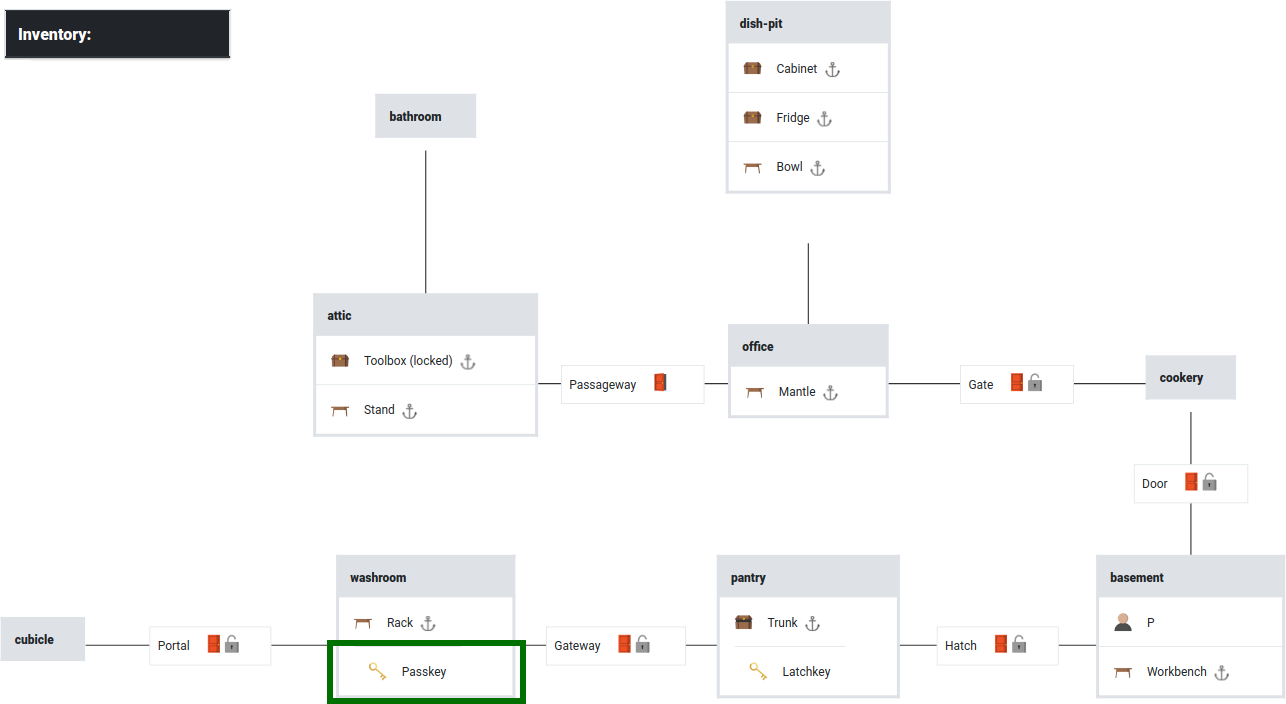}
            \caption{A generated Treasure Hunter game with difficulty 20. The green rectangle identifies the quest object, a \entity{passkey}, which can be found on a \entity{rack} in the \entity{washroom}. The player, \entity{P}, is currently two rooms away in the \entity{basement} separated by two closed doors.}
            \label{fig:th_level20}
        \end{figure}


\section{Current Limitations}
  \label{sect:future}

  \textbf{Complex Quest Generation} We define a quest as a sequence of actions where each action depends of the outcomes of its predecessor (Section~\ref{sect:generation_quest}). This limits us to simple quests where each action is based on only the immediately previous action.
  Quests are rarely so straightforward in text adventure games, often being composed of multiple sub-quests.
  One method for generating more complex quests would be by treating a quest as a directed graph of dependent actions rather than a linear chain.

  \textbf{Time-based Events} In the current implementation, there is no support for triggering state changes without corresponding user actions. This would enable doors that automatically lock after a certain number of player steps or traps that trigger periodically.

  \textbf{Non-Player Characters (NPCs)} NPCs are characters that are not controlled by the player and are capable of performing autonomous or reactive actions that alter the game state.
  NPC interaction is quite common in several text-based games, but is not currently supported by the game generator.

  \textbf{Multi-User Dungeon(MUD)} Some text-based games allow multiple users to interact with the world at the same time. Users may need to cooperate with each other or to compete against each other, depending on the game. The framework currently supports only single-agent games.

  \textbf{Text Generation}
  Using a CFG for text generation makes it difficult to ensure continuity between sentences.

\section{Conclusion}
    \label{sect:conc}
    We introduced TextWorld, a sandbox learning environment for the training and evaluation of RL agents on text-based games.
    After surveying the machine-learning challenges of text-based games, we cast them in the formalism of reinforcement learning.
    We described how the generative mechanisms of TextWorld can be used to work up towards the full complexity of hand-authored games and introduced a preliminary set of benchmark games for this purpose.
    We evaluated several baseline agents on this set and a curated list of hand-authored games that we analyzed.

    In future work, we will develop more complex benchmark tasks and investigate novel approaches to representation learning, language understanding and generation, and RL for text-based games.
\section{Acknowledgement}
  We thank Adam Ferguson and Marion Zepf for many helpful discussions on game analysis. We thank Alessandro Sordoni, Greg Yang, Philip Bachman, Ricky Loynd, Samira Ebrahimi Kahou, and Yoshua Bengio for helpful suggestions and comments on research perspectives.

\clearpage
\bibliographystyle{plainnat}
\bibliography{biblio}

\appendix
\clearpage
\section{Typical Text-based Obstacles}
\label{sect:appendix_puzzle_types_long}

 \begin{description}
     \item[Language] Language presents a number of difficulties Parser complexity (what a parser can and cannot handle), distracting in-game descriptions, sarcasm (e.g., If the game responds \literal{Yeah right, that's totally going to work} when you enter a command, a solid understanding of sarcasm would prevent you from repeating the command), and fictional languages.
      \item[Maze] One of the oldest text adventure obstacles, the most basic maze is a grid of rooms with similar names and/or descriptions. The player has to find the right path through the maze to reach a goal. Many text-based games take care to innovate/complicate mazes. Some innovations include:
        \begin{itemize}
            \item Mazes whose solution is determined stochastically, either on a play-through by play-through basis or on a turn by turn basis.
            \item 	Mazes whose configuration changes due to a player action, as in Zork III~\citep{if:ZorkIII}, where the player can move some walls in the maze in order to solve it.
            \item Mazes that are not solvable by navigation: In Photopia, for example, the player is wearing a spacesuit while navigating the maze. If the player removes the suit, it is revealed that the player has a pair of wings. The player is then able to fly over the maze.
        \end{itemize}
      \item[Clue Hunting] Detective games require the player to hunt for clues that will tell them where to go next. In Sherlock, the game begins when the player receives a coded poem that directs them to Westminster Abbey, where they have to solve a number of riddle puzzles in order to find out where to find five different gems, each of which provides the player with further clues. A point of interest here is whether or not the player actually needs to get the hints in order to win the game, or whether the player could win the game if they already knew the contents of the hints. Sherlock~\citep{if:sherlock} is an example of the former – some hints trigger crucial game state changes only once the player examines them. Other games mix state-changing hints with purely informative ones. For example, in Inhumane~\citep{if:Inhumane}, a clue gives you directions through a maze which is still solvable if you don’t find the clue. However, the end of the maze yields another clue. When you read this second clue, an important portion of the map becomes available to the player.
      \item[Treasure Hunting] Some treasures double as clues (as is the case with the gems in Sherlock) while other treasures are necessary in order to unlock further treasures. In Infidel, for example, you are given the opportunity to take some gem clusters early in the game. Towards the end of the game, these clusters are used to unlock a treasure chest.~\citep{if:infidel}
      \item[Trivia] Many games base their puzzles off of trivia/world knowledge, which the player might need to know in order to solve a puzzle. Many puzzles in the Hitchhiker's Guide to the Galaxy game are easier to solve if you are familiar with the source material. For example, if you've read the book, you'd already know that a Babel Fish is able to decode any language in the universe. In the game, then, you'll know what to look for when confronted with an alien language~\citep{if:Hhgg}. In The Enterprise Incidents, the player must solve a word puzzle whose solution, the word \emph{firefly}, is not mentioned elsewhere in the game.~\citep{if:Enter} In Goldilocks is a FOX, knowledge of fairy tales is required to solve puzzles involving magic beans, bears, and porridge.~\citep{if:Gold} Sherlock contains intricate riddle poems whose solutions are allusions to famous historical figures buried in Westminster Abbey. The player is also required decode a riddle poem and realize that references to ``the conquest'' and ``the fire'' refer to the Battle of Hastings and the Great Fire of London, respectively. Finally, the player is expected to know the dates for these two dates in order to subtract one from the other in order to solve a puzzle.~\citep{if:sherlock} A number of modern text-based games (Curses, OMNIQUEST, All Quiet on the Library Front, Inhumane, and the later Zork games, to name a few) play off of established text adventure tropes, clich\'{e}s, and catchphrases (common passwords across games include \literal{XYZZY} and \literal{Plugh} from Cave Adventure~\citep{if:Advent}, and magic spells learned in the Enchanter series carry over from game to game, although they have to be `learned' in-game in order to be used).
      \item[Self-Maintenance] Hunger, thirst, fatigue are the most common survival elements in text adventures. Players might have to keep space open in their inventory for food and drink, and will be prompted from time to time that they are getting hungry or thirsty, or that they might need to rest. Similar obstacles that pop up are: hypothermia (when swimming in Zork III~\citep{if:ZorkIII}), the classic lamp from Zork~\citep{if:Zork} (which reappears and is innovated on in the sequels, where it can go out if you neglect to shake it or turn it back on), and torches from Infidel (which you are required to regularly dip in oil and light)~\citep{if:infidel}. One complication of self-maintenance is whether the action is one time or continuous (in Infidel, for example, you need to gather food and drink, but once they’re gathered, you have an unlimited amount).
      \item[Combat] Combat can come in a variety of forms. At its most basic level, a player will be confronted with an enemy and can thwart it by typing \cmd{kill enemy}. To complicate things further, the player might not be equipped to kill the enemy with their bare hands, and must first acquire a weapon before \cmd{kill enemy} will return \literal{enemy killed}. One level higher, the player might have to \cmd{attack enemy} and \cmd{dodge enemy} or \cmd{aim at X} (where X = a weak spot you learn about elsewhere in the game) before killing the enemy. Higher levels include stochastic combat (where attacks may or may not land), optional combat (where the only solution to the fight is to avoid it) and puzzle combat. In Reverberations, for example, the player encounters an enemy in the cosmetics section of a department store. To win the fight, the player must spray the enemy with some nearby perfume.~\citep{if:Reverb}
      \item[Time] In Wishbringer, the player has a limited amount of time at the start of the game to deliver a note. The player must use this time to collect items and information that will be inaccessible once they've delivered the note.~\citep{if:Wishbringer} In Sherlock, the player is given 48 hours of in-game time to solve the mystery. Certain locations and events are only accessible at certain times.~\citep{if:sherlock}
      \item[Mechanical Puzzle] Some examples can be found in Infidel and Inhumane, where the player has to move objects around a room, leave objects in rooms, and prop open doors with wooden beams in order to avoid being caught in traps.~\citep{if:infidel}~\citep{if:Inhumane}
      \item[Persistence] In some games, Anchorhead for example, you must repeat \cmd{talk to} commands in order to get more information out of an NPC.~\citep{if:Anchor}
      \item[Stochasticity] Some examples of stochasticity include randomly generated maze configurations, randomly decided combat action outcomes, and randomly selected code-words for puzzles. Text-based game players deal with stochasticity with the \cmd{UNDO} action or by saving often and especially right before taking any kind of risky action.
      \item[Cruelty Scale] The Cruelty Scale was designed by text-based game author Andrew Plotkin as a broad method of determining the difficulty of text-based games. The main element considered by the scale is whether or not the game can be made impossible to win, and how obvious (or not) this is made to the player. The scale has five ranks: Merciful games are impossible to die or get stuck in, Polite games warn you before you do something that can kill or trap you or otherwise render the game unwinnable. Tough games can kill the player or be rendered unwinnable, but they'll warn the player before this happens. Nasty games can be rendered unwinnable without warning, but there will be an indication that the game is now unwinnable after the fact. Cruel games can be rendered unwinnable, and there's no warning beforehand and no indication afterwards.
      \item[Length] Text-based games vary wildly in length, from the solvable-in-twenty-moves inform demo Acorncourt to the 1000+ moves required to complete Anchorhead, text-based games have many elements that impact their length. These include: Number of possible commands, number of rooms, length of descriptive text, (Reverberations, for example, is solvable with a relatively low number of commands (roughly 50), but features long descriptive passages for almost every room and command) side quests and multiple endings, and game difficulty. However, text-based game review sites and discussion boards often divide games up into short demo type games, medium uncomplicated games, and long difficult games.
    \end{description}
\clearpage
\begin{landscape}
\section{Curated List of text-based Games}
\label{sect:appendix_curated_list}

\begin{small}
\begin{longtable}{lcccccccccccc}
  \toprule
  Game & \# Rooms &  Max  & Maze & Trivia & Self- & Combat & Time & Mechanical & Persistence & Stochasticity & Forgiveness\\
   &  &  Score  &  &  & maintenance &  &  & Puzzle & & & \\
  \midrule
  \endhead  
  \href{http://ifdb.tads.org/viewgame?id=tqvambr6vowym20v}{The Acorn Court} & 1 & 30 & No & No & No & No & No & Yes & No & No & Merciful \\
  \href{http://ifdb.tads.org/viewgame?id=fft6pu91j85y4acv}{Adventure} & 30+ & 350 & Yes & Yes & No & Yes & Yes & Yes & Yes & No & Tough \\
  \href{http://ifdb.tads.org/viewgame?id=dy4ok8sdlut6ddj7}{Adventureland} & 30+ & 100 & Yes & Yes & No & Yes & Yes & Yes & Yes & Yes & Tough \\
  \href{http://ifdb.tads.org/viewgame?id=400zakqderzjnu1i}{All Quiet on the Library Front} & 2+ & 30 & No & Yes & No & No & No & No & No & No & Merciful \\
  \href{http://ifdb.tads.org/viewgame?id=op0uw1gn1tjqmjt7}{Anchorhead} & 30+ & 100 & No & Yes & No & Yes & Yes & No & Yes & No & Cruel \\
  \href{http://ifdb.tads.org/viewgame?id=rwseuddvj1gbo481}{The Awakening} & 30+	& 50 & No & Yes & No & Yes & Yes& Yes & Yes & No & Polite \\
  \href{http://ifdb.tads.org/viewgame?id=x6ne0bbd2oqm6h3a}{Balances} & 10+ & 51 & No & Yes & No & No & No & Yes & No & No & Polite \\
  \href{http://ifdb.tads.org/viewgame?id=b0i6bx7g4rkrekgg}{Ballyhoo} & 30+ & 300  & Yes  & Yes  & Yes & No & Yes & Yes & No & Yes & Tough \\
  \href{http://ifdb.tads.org/viewgame?id=plvzam05bmz3enh8}{Curses} & 30+  & 550  & No  & Yes  & No  & No  & Yes  & No  & Yes  & Yes  & Polite \\
  \href{http://ifdb.tads.org/viewgame?id=4ao65o1u0xuvj8jf}{Cutthroat} & 30+  & 250  & Yes  & No  & Yes  & Yes  & Yes  & Yes  & Yes  & Yes  & Polite \\
  \href{http://ifdb.tads.org/viewgame?id=x85otcikhwp8bwup}{Deephome} & 10+  & 300 & Yes  & No  & No  & Yes  & Yes  & Yes  & Yes & No & Polite \\
  \href{http://ifdb.tads.org/viewgame?id=1po9rgq2xssupefw}{Detective} & 10+ & 360 & No & No & No & Yes & No & No & No & No & Cruel \\
  \href{http://ifdb.tads.org/viewgame?id=sjiyffz8n5patu8l}{Dragon Adventure} & 2+  & 25  & No  & No  & No  & Yes  & No  & No  & No  & No  & Merciful \\
  \href{http://ifdb.tads.org/viewgame?id=vu4xhul3abknifcr}{Enchanter} & 30+ & 400 & Yes & No & Yes & Yes & Yes & Yes & Yes & Yes & Cruel \\
  \href{http://ifdb.tads.org/viewgame?id=ld1f3t5epeagilfz}{The Enterprise Incidents} & 10 & 10 & No & Yes & No & No & Yes & No & Yes & No & Merciful \\
  \href{http://ifdb.tads.org/viewgame?id=59ztsy9p01avd6wp}{Goldilocks is a FOX!} & 30+ & 100 & Yes & Yes & No & No & No & Yes & No & No & Polite \\
  \href{http://ifdb.tads.org/viewgame?id=ouv80gvsl32xlion}{The Hitchhiker's Guide to the Galaxy} & 30+ & 100 & Yes & Yes & Yes & No & Yes & Yes & Yes & Yes & Cruel \\
  \href{http://ifdb.tads.org/viewgame?id=jnfkbgdgopwfqist}{Hollywood Hijnx} & 10+  & 150  & Yes  & Yes  & No  & Yes  & No  & Yes  & No  & No  & Nasty \\
  \href{http://ifdb.tads.org/viewgame?id=anu79a4n1jedg5mm}{Infidel} & 10+ & 400 & Yes & Yes & Yes & No & No & Yes & No & No & Cruel \\
  \href{http://ifdb.tads.org/viewgame?id=wvs2vmbigm9unlpd}{Inhumane} & 10+ & 400 & Yes & Yes & No & No & No & Yes & No & No & Polite \\
  \href{http://ifdb.tads.org/viewgame?id=hu60gp1bgkhlo5yx}{The Jewel of Knowledge} & 30+  & 90  & Yes  & No  & No  & Yes  & No  & Yes  & Yes  & Yes  & Tough \\
  \href{http://ifdb.tads.org/viewgame?id=3p9fdt4fxr2goctw}{Leather Goddesses of Phobos} & 30+  & 316  & Yes  & Yes  & No  & Yes  & Yes  & Yes  & Yes  & Yes  & Cruel \\
  \href{http://ifdb.tads.org/viewgame?id=4wd3lyaxi4thp8qi}{Mother Loose} & 10+ & 50 & No & Yes & No & No & No & No & No & No & Merciful \\
  \href{http://ifdb.tads.org/viewgame?id=mohwfk47yjzii14w}{Lost Pig} & 2+ & 7 & No & No & No & No & No & Yes & No & No & Merciful \\
  \href{http://ifdb.tads.org/viewgame?id=r6g7pflngn3uxbam}{The Ludicorp Mystery}  & 30+  & 150  & Yes  & No  & No  & No  & No  & Yes  & No  & No  & Polite \\
  \href{http://ifdb.tads.org/viewgame?id=jhbd0kja1t57uop}{The Lurking Horror}  & 10+  & 100  & Yes  & No  & Yes  & Yes  & Yes  & Yes  & Yes  & No & Polite \\
  \href{http://ifdb.tads.org/viewgame?id=273o81yvg64m4pkz}{The Meteor, the Stone and a Long Glass of Sherbet} & 30+ & 30 & No & Yes& No & Yes & Yes & Yes & No & Yes & Cruel\\
  \href{http://ifdb.tads.org/viewgame?id=q36lh5np0q9nak28}{Monsters of Murdac} & 30+ & 250 & Yes & No & No & Yes & Yes & No & No & Yes & Nasty\\
  \href{http://ifdb.tads.org/viewgame?id=ydhwa11st460g9u3}{Night at the Computer Center} & 10+ & 10 & Yes & Yes & No & No & No & No & No & No & Merciful \\
  \href{http://ifdb.tads.org/viewgame?id=mygqz9tzxqvryead}{OMNIquest} & 2+ & 50 & No & Yes & No & Yes & No & No & No & No & Polite \\
  \href{http://ifdb.tads.org/viewgame?id=llchvog0ukwrphih}{Pentari} & 2+ & 70 & No & No & No & Yes & Yes & No & No & Yes & Polite \\
  \href{http://ifdb.tads.org/viewgame?id=xe6kb3cuqwie2q38}{Planetfall} & 10+ & 80 & No & No & Yes & Yes & Yes & Yes & Yes & Yes & Polite \\
  \href{http://ifdb.tads.org/viewgame?id=ddagftras22bnz8h}{Plundered Hearts} & 10+ & 25 & No & Yes & No & Yes & Yes & Yes & No & No & Polite\\
  \href{http://ifdb.tads.org/viewgame?id=bx8118ggp6j7nslo}{Return to Karn} & 30+ & 170 & No	& Yes & No & No & Yes & Yes & Yes & Yes & Nasty\\
  \href{http://ifdb.tads.org/viewgame?id=dop7nbjl90r5zmf9}{Reverberations} & 10+ & 50 & No & Yes & No & Yes & Yes & Yes & No & No & Polite \\
  \href{http://ifdb.tads.org/viewgame?id=56wb8hflec2isvzm}{Seastalker} & 30+ & 100 & No & No & No & Yes & Yes & Yes & Yes & No & Nasty\\
  \href{http://ifdb.tads.org/viewgame?id=j8lmspy4iz73mx26}{Sherlock: The Riddle of the Crown Jewels} & 30+ & 100 & No & Yes & No & Yes & Yes & Yes & Yes & Yes & Nasty \\
  \href{http://ifdb.tads.org/viewgame?id=lidg5nx9ig0bwk55}{Sorcerer} & 10+ & 400 & Yes & Yes & Yes & Yes & Yes & Yes & Yes & Yes & Nasty \\
  \href{http://ifdb.tads.org/viewgame?id=wqsmrahzozosu3r}{Spellbreaker} & 30+ & 600 & Yes & Yes & No & Yes & Yes & Yes & Yes & Yes & Cruel \\
  \href{http://ifdb.tads.org/viewgame?id=tqpowvmdoemtooqf}{Spiritwrak} & 30+ & 250 & Yes & Yes & Yes & Yes & Yes & Yes & Yes & Yes & Polite \\
  \href{http://ifdb.tads.org/viewgame?id=kq9qgjkf2k6xn1c0}{The Temple} & 10+ & 35 & No & Yes & No & Yes & Yes & No & Yes & No & Polite \\
  \href{http://ifdb.tads.org/viewgame?id=bv8of8y9xeo7307g}{Theatre} & 30+ & 50 & No & Yes & No & Yes & Yes & Yes & Yes & No & Polite \\
  \href{http://ifdb.tads.org/viewgame?id=j18kjz80hxjtyayw}{Trinity} & 30+ & 100 & Yes & Yes & No & No & Yes & Yes & No & Yes & Tough \\
  \href{http://ifdb.tads.org/viewgame?id=ic0ebhbi70bdmyc2}{Tryst of Fate} & 30+ & 350 & Yes & Yes & No & Yes & Yes & Yes & Yes & Yes & Polite \\
  \href{http://ifdb.tads.org/viewgame?id=z02joykzh66wfhcl}{Wishbringer} & 30+ & 100 & Yes & No & No & Yes & Yes & Yes & Yes & Yes & Tough \\
  \href{http://ifdb.tads.org/viewgame?id=rw7zv98mifbr3335}{Zenon} & 10+ & 20 & No & No & No & No & Yes & No & No & No & Polite \\
  \href{http://ifdb.tads.org/viewgame?id=0dbnusxunq7fw5ro}{Zork I} & 30+ & 350 & Yes & Yes & Yes & Yes & Yes & Yes & Yes & Yes & Cruel \\
  \href{http://ifdb.tads.org/viewgame?id=yzzm4puxyjakk8c4}{Zork II} & 30+ & 400 & Yes & Yes & Yes & Yes & Yes & Yes & No & No & Cruel \\
  \href{http://ifdb.tads.org/viewgame?id=vrsot1zgy1wfcdru}{Zork III} & 30+ & 7 & Yes & Yes & Yes & Yes & Yes & Yes & Yes & Yes & Cruel \\
  \href{http://ifdb.tads.org/viewgame?id=40hswtkhap88gzvn}{Zork: The Undiscovered Underground}  & 10+ & 100 & Yes & Yes & No & No & Yes& Yes & Yes & Yes & Tough\\
  \bottomrule
  \caption[Curated list]{Information we collected during our analysis of text-based games.}
\label{tab:curated_list}
\end{longtable}
\end{small}

\end{landscape}
\clearpage

\subsection{Game Notes}
  \label{sect:additional_notes_on_games}
  Below are observations for some of the games we analyzed.
  \begin{description}

      \item[\href{http://ifdb.tads.org/viewgame?id=tqvambr6vowym20v}{The Acorn Court}]
        Written as an Inform demo. Very short game with only one room. Contains one multi-part spatial puzzle.

      \item[\href{http://ifdb.tads.org/viewgame?id=fft6pu91j85y4acv}{Adventure}]
        First IF. Difficult, stochastically generated mazes, introduces made up words. We used the recompiled with Inform 6 (release 9) version which starts with an initial score of 36 points.

      \item[\href{http://ifdb.tads.org/viewgame?id=op0uw1gn1tjqmjt7}{Anchorhead}]
        Heavily text based. Requires a combination of world knowledge and knowledge gleaned from in-game texts. When speaking with townspeople, you have to occasionally be persistent. Awareness of Lovecraftian cliches is helpful.

      \item[\href{http://ifdb.tads.org/viewgame?id=x6ne0bbd2oqm6h3a}{Balances}]
        Another Inform demo, features made up words and spells that the player must remember. Rare words used in the context of magic and fantasy are also used.

        \item[\href{http://ifdb.tads.org/viewgame?id=x85otcikhwp8bwup}{Deephome}]
        Medium length game with an in game library the player can consult in order to figure out the solutions to puzzles.

      \item[\href{http://ifdb.tads.org/viewgame?id=sjiyffz8n5patu8l}{Dragon Adventure}]
        Short game written to introduce children to text based games.

      \item[\href{http://ifdb.tads.org/viewgame?id=1po9rgq2xssupefw}{Detective}]
        Poorly written game. Navigation isn't very logical (you can walk east into a room whose only exit is to the north). Gameplay is very simple - you only need to use navigation commands to reach the end of the game.

      \item[\href{http://ifdb.tads.org/viewgame?id=vu4xhul3abknifcr}{Enchanter}]
        Long game with a complex spell-casting system that requires player to memorize and record spells.


      \item[\href{http://ifdb.tads.org/viewgame?id=59ztsy9p01avd6wp}{Goldilocks is a FOX!}]
        Requires trivia knowledge (you need to know about Jack and the Beanstalk in order to know what to do when someone offers you magic beans). Casual language with lots of pop cultural references.

      \item[\href{http://ifdb.tads.org/viewgame?id=ouv80gvsl32xlion}{Hitchhiker's Guide to the Galaxy}]
        Incredibly difficult to play. Several stochastic obstacles (A maze without a set solution, An object from a subset of objects is randomly made into a crucial object). Features many irrational objects and events, including objects showing up in your inventory without warning. Leans on world knowledge a bit as well, as when the navigation controls change to Port, Starboard, Aft, and Bow/Fore. Some puzzles are solvable if you are familiar with the books. Language is also sarcastic and, at times, deliberately misleads the player.


      \item[\href{http://ifdb.tads.org/viewgame?id=anu79a4n1jedg5mm}{Infidel}]
        Many spatial puzzles. Draws on world knowledge (player must decode hieroglyphics and pictograms). Instruction manual contains the navigation instructions for finding the pyramid. Contains some potential one-way paths (you need to tie a rope to an object in one room in order to re-enter it).

      \item[\href{http://ifdb.tads.org/viewgame?id=wvs2vmbigm9unlpd}{Inhumane}]
        Parody of infidel but with a more straightforward, simpler map. Has a meta game where you try to nearly solve puzzles and then kill yourself. Some familiarity with Infidel makes playing it easier. Contains some clue-hunting that doubles as room unlocking, you can also find directions through a maze in the game. Contains a gag ending triggered by entering \cmd{s} instead of \cmd{south} or vice versa as your final command. There are mazes, but there are also some unconnected rooms that have identical names (T-Intersections appear in different areas).

      \item[\href{http://ifdb.tads.org/viewgame?id=400zakqderzjnu1i}{All Quiet on the Library Front}]
        Relatively simple treasure hunt with minimal side-quests and Easter eggs. Contains some text-based game in-jokes that do not impact the gameplay, but might be distracting.

      \item[\href{http://ifdb.tads.org/viewgame?id=mohwfk47yjzii14w}{Lost pig}]
        Game text is written in “cave-person speak”. Otherwise a simple pig hunt with some magic spell based puzzles.

      \item[\href{http://ifdb.tads.org/viewgame?id=jhbd0kja1t57uop}{The Lurking Horror}]
        Medium length game that requires you to understand which rooms are above which other rooms in order to guess the solution to a puzzle.



      \item[\href{http://ifdb.tads.org/viewgame?id=tqpowvmdoemtooqf}{Spiritwrak}]
        Features a massive map with a working subway system. Uses an Enchanter-style spell system.

     \item[\href{http://ifdb.tads.org/viewgame?id=j18kjz80hxjtyayw}{Trinity}]
        Long game with surreal imagery. Map made of several sub-worlds linked through a central hub.

      \item[\href{http://ifdb.tads.org/viewgame?id=j8lmspy4iz73mx26}{Sherlock: The Riddle of the Crown Jewels}]
        Very dependent on world knowledge. Time based. Some word puzzles. Player can easily get stuck in an unwinnable state without knowing it, and is expected to have a familiarity with Victorian England and British history and culture and general.

      \item[\href{http://ifdb.tads.org/viewgame?id=z02joykzh66wfhcl}{Wishbringer}]
        Designed for beginner players. Features a hint system and an object that grants the player "wishes" that help them bypass puzzles.


      \item[\href{http://ifdb.tads.org/viewgame?id=0dbnusxunq7fw5ro}{Zork I}]
        Classic text-based game treasure hunt with dungeon theme. In addition to the treasure hunt component, player has to contend with a thief, combat with a troll, and navigate a maze. Knowledge about \textit{The Odyssey} can help the player defeat a Cyclops.

      \item[\href{http://ifdb.tads.org/viewgame?id=yzzm4puxyjakk8c4}{Zork II}]
        Sequel to Zork I. Contains a maze modelled after a baseball diamond that confused some European players when first released.

      \item[\href{http://ifdb.tads.org/viewgame?id=vrsot1zgy1wfcdru}{Zork III}]
        Sequel to Zork II. High score can be achieved without winning the game. Game rewards the player for finding innovative solutions to problems rather than for solving puzzles, navigating to rooms, or retrieving objects.

  \end{description}

\section{Inform 7}
  \label{sect:appendix_inform7}
  Inform refers to a domain-specific programming language and additional tooling for interactive fiction (\ie text-based games). It is regarded as the most natural language-like programming language. It was originally created in 1993 by Graham Nelson and he later released Inform 7 (briefly known as Natural Inform). We decide to use Inform7 so we could leverage Inform's impressive parser that benefited from more than two decades of tweaks/fixes. Inform 7 source compiles to Inform 6 source, a weakly-typed multiple-inheritance traditional programming language, before compiling to Z or glulx code. See \url{http://inform7.com/} for more information on Inform 7. Also, here is a supplemental resource for some of the technical details: \url{http://www.ifwiki.org/index.php/Inform_7_for_Programmers}.

\clearpage

\end{document}